\documentclass{article}

\PassOptionsToPackage{numbers}{natbib}

\usepackage[preprint]{neurips_2025}
\usepackage{mathtools}
\usepackage{multicol}
\usepackage{wrapfig}
\usepackage{enumitem}

\usepackage[utf8]{inputenc} 
\usepackage[T1]{fontenc}    
\usepackage[hidelinks]{hyperref}       
\usepackage{url}            
\usepackage{booktabs}       
\usepackage{amsmath, amsfonts, amssymb, amsthm}       
\usepackage{nicefrac}       
\usepackage{microtype}      
\usepackage[dvipsnames]{xcolor}         

\usepackage{tikz-network}
\usepackage{csquotes}
\usepackage{comment}
\usepackage{graphicx} 
\usepackage[ruled, linesnumberedhidden, vlined]{algorithm2e}
\usepackage{algorithmic}
\usepackage{subcaption}
\usepackage{changepage}

\newtheorem{definition}{Definition}[section]

\newif\ifdraft
\drafttrue  

\ifdraft
  \newcommand{\mansi}[1]{{\color{ForestGreen} \textbf{Mansi:} \enquote{#1}}}
  \newcommand{\nathaniel}[1]{{\color{Blue} \textbf{Nathaniel:} \enquote{#1}}}
  \newcommand{\kyle}[1]{{\color{CadetBlue} \textbf{Kyle:} \enquote{#1}}}
  \newcommand{\tian}[1]{{\color{red} \textbf{Tian:} \enquote{#1}}}
  \newcommand{\ian}[1]{{\color{BurntOrange} \textbf{Ian:} \enquote{#1}}}
  \newcommand{\msakarvadia}[1]{{\color{ForestGreen} [#1 -MS]}}
  \newcommand{\nhudson}[1]{{\color{Blue} [#1 -NH]}}
  \newcommand{\kchard}[1]{{\color{CadetBlue} [#1 -KC]}}
  \newcommand{\ifoster}[1]{{\color{BurntOrange} [#1 -IF]}}
  \newcommand{\tli}[1]{{\color{Red} [#1 -TL]}}
\else
  \newcommand{\mansi}[1]{}
  \newcommand{\nathaniel}[1]{}
  \newcommand{\kyle}[1]{}
  \newcommand{\ian}[1]{}
  \newcommand{\tian}[1]{}
  \newcommand{\msakarvadia}[1]{}
  \newcommand{\aajith}[1]{}
  \newcommand{\nhudson}[1]{}
  \newcommand{\kchard}[1]{}
  \newcommand{\ifoster}[1]{}
  \newcommand{\tli}[1]{}
\fi

\newcommand{\barabasialbert}{\textbf{Barab\'{a}si-Albert}}
\newcommand{\stochasticblock}{\textbf{Stochastic Block}}
\newcommand{\wattsstrogatz}{\textbf{Watts-Strogatz}}

\newcommand{\BA}{\textbf{BA}}
\newcommand{\SB}{\textbf{SB}}
\newcommand{\WS}{\textbf{WS}}

\newcommand{\fl}{\texttt{FL}}
\newcommand{\unweighted}{\texttt{Unweighted}}
\newcommand{\weighted}{\texttt{Weighted}}
\newcommand{\random}{\texttt{Random}}
\newcommand{\degree}{\texttt{Degree}}
\newcommand{\betweenness}{\texttt{Betweenness}}
\newcommand{\neighbors}{\textit{neighbors}}

\DeclareMathOperator{\Dir}{Dir}

\def\equationautorefname~#1\null{Eq~(#1)\null}

\title{

    Topology-Aware Knowledge Propagation \\ in Decentralized Learning
}

\author{%
    Mansi Sakarvadia\textsuperscript{\textnormal{1}}\thanks{Correspondence to \href{mailto:sakarvadia@uchicago.edu}{\texttt{sakarvadia@uchicago.edu}}}~~,
    Nathaniel Hudson\textsuperscript{\textnormal{1,2}},
    Tian Li\textsuperscript{\textnormal{1}},
    {\bf Ian Foster}\textsuperscript{1,2},
    {\bf Kyle Chard}\textsuperscript{1,2}\\
     \textsuperscript{1}University of Chicago,
     \textsuperscript{2}Argonne National Laboratory 
}
\usepackage[framemethod=TikZ]{mdframed}

\begin{document}

\maketitle

\begin{abstract}
    Decentralized learning enables collaborative training of models across naturally distributed data without
    centralized coordination or maintenance of a \textit{global} model. Instead, devices are organized in arbitrary communication topologies, in which they can only communicate with neighboring devices. Each device maintains its own local model by training on its local data and integrating new knowledge via model aggregation with neighbors.
    Therefore, knowledge is propagated across the topology via successive aggregation rounds.
    We study, in particular, the propagation of out-of-distribution (OOD) knowledge. 
    We find that popular decentralized learning algorithms struggle to propagate OOD knowledge effectively to all devices. 
    Further, we find that both the location of OOD data within a topology, and the topology itself, significantly impact OOD knowledge propagation. 
    We then propose \textbf{topology-aware aggregation} strategies to accelerate (OOD) knowledge propagation across devices.
    These strategies improve OOD data accuracy, compared to topology-unaware baselines, by 123\% on average across models in a topology\footnote{\url{https://github.com/msakarvadia/topology_aware_learning}}.
\end{abstract}

\section{Introduction}
\label{sec:introduction}
Most machine learning training data are
generated, collected, and sensed from decentralized sources:
Internet-of-Things~\cite{nguyen2021federated}, edge/fog/cloudlet computing systems~\cite{zhou2019edge, wang2019adaptive, zhou2020privacy, withana2023ckn}, sensor networks~\cite{sage}, smart grids~\cite{hudson2021framework, molokomme2022edge}, and smart transportation networks~\cite{zhou2020drle, hudson2022smart}.
Because most data are naturally decentralized, a question arises: \textit{How do we train models across decentralized data?}

A common solution is \textit{centralized learning}, in which decentralized data are sent to a central location where training occurs~\cite{krizhevsky2014one}. However, centralized learning incurs data transfer costs~\cite{hudson2021framework, hudson2022smart} and raises data privacy concerns~\cite{kairouz2021advances, zhang2021survey, li2020review}. 
\textit{Federated learning}~(FL) addresses these concerns
by training local models directly on devices located at each data generation site and periodically aggregating these local models into a single global model at a central server~\cite{mcmahan2017communication}. 
While FL has been widely adopted~\cite{wen22survey}, its rigid client-server model is both a single point of failure and is ill-suited for 
many real-world distributed systems (e.g., ad hoc networks) \cite{zheng2023federated, kang2020reliable}.  
\textit{Decentralized learning} is an alternative which addresses the data transfer and privacy concerns of centralized learning and accommodates arbitrary and unstable communication topologies found in many wide area networks which are a limiting factor in federated learning~\cite{koloskova2019decentralized_icml, koloskova2020unified}.

Decentralized learning enables collaborative learning across devices without creating a single global model or requiring that data be centralized~\cite{beltran2023decentralized, hegedHus2019gossip}. 
Instead, each device maintains its model by training over local data and integrating additional (non-local) knowledge by periodically receiving neighboring devices' models and aggregating them with its local model.
Devices are organized in a flexible
\textit{topology} in which nodes represent devices and edges/links represent communication channels. 
Devices can be located at data generation sites, with communication channels between devices dependent on factors like physical locality, administrative connections, and privacy concerns \cite{edge2019zhao}.
However, this flexibility can come at a cost, in particular slower convergence and regional/hyper-personalized models that lack knowledge from distant devices. To prevent this from happening, we aim for each device-specific model to be performant over the global data distribution across all devices in a topology;
device-specific models must be generalizable beyond their local data distribution
so that they are performant on \textit{out-of-distribution}~(OOD) inference requests. 
This is especially challenging in decentralized learning as
the only way for device-specific knowledge to propagate 
in a topology 
is by \enquote{hopping} between devices via successive aggregation rounds.

\begin{figure*}[t]
\centering
    \begin{subfigure}[b]{0.5\textwidth}
        \centering
        \includegraphics[width=\linewidth]{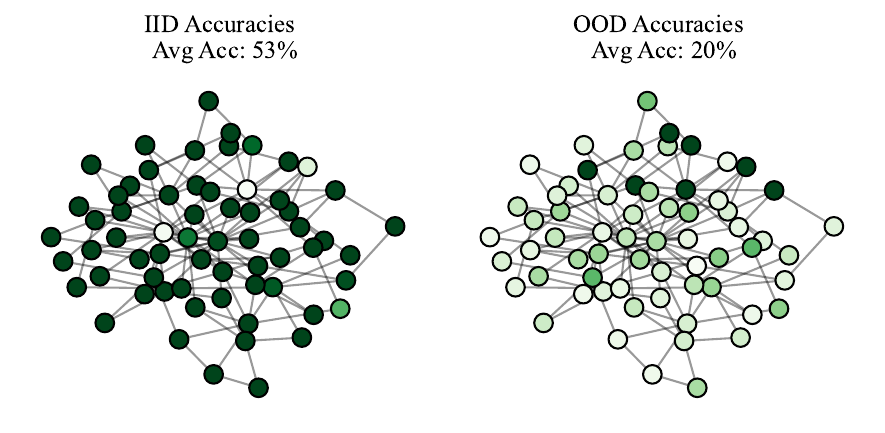}
        \caption{Baseline: Topology-Unaware}
        \label{fig:weighted_bd}
    \end{subfigure}%
    \vrule
    \begin{subfigure}[b]{0.5\textwidth}
        \centering
        \includegraphics[width=\linewidth]{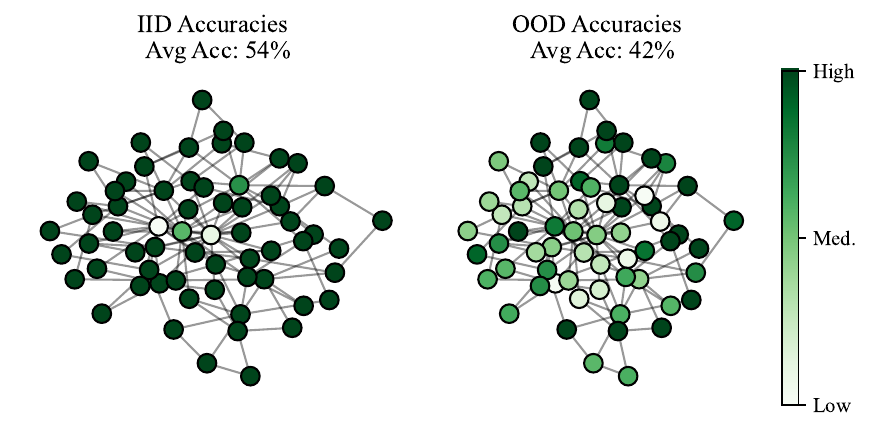}
        \caption{Proposed: Topology-Aware}
        \label{fig:degree_bd}
    \end{subfigure}%
    \caption{\textbf{Topology-(un)aware aggregation for IID vs.\ OOD knowledge propagation.}
    CIFAR10 is distributed across 64 nodes: OOD data placed on node with the fourth highest degree. 
    Aggregation strategy for topology-unaware is \unweighted{} and topology-aware is \degree{}. 
    \textbf{Green} indicates \textit{higher} test accuracy on the respective dataset after 40 rounds of training; \textbf{white} indicates the opposite.
    Our proposed topology-aware method (right) achieves higher test OOD accuracies without sacrificing IID accuracies. 
    }
    \label{fig:topo_v_non_topo_aware}
\end{figure*}

Here, we study knowledge propagation in decentralized topologies by asking: \textit{
How can each device-specific model learn from \textbf{all} data present in a topology, regardless of its location
, in as few aggregation rounds as possible?
}
This goal is especially challenging in settings where data are not \textit{independently and identically distributed (IID)} across devices
as devices have no knowledge of how data are distributed globally.
We study the extreme case in which most data in a topology are IID, 
with the exception of a single device which contains OOD data. 
We find that 
existing decentralized learning strategies~\cite{koloskova2020unified} struggle to propagate the OOD knowledge. 
Further, we find that both the topology and location of data within the topology impacts OOD knowledge propagation---devices in non-central locations, for example, exhibit poor knowledge propagation. 

To address the variability of knowledge propagation due to 
topology, \textbf{we propose topology-aware aggregation strategies for decentralized learning}. Traditional aggregation strategies
fail to account for a node's 
(non)beneficial 
location in a topology. Topology-aware aggregation strategies instead allow each device to account for its own and its neighbors' location in a topology when aggregating models. 
For example, devices with many neighbors are well positioned to act as information hubs: they can both ingest and disperse knowledge to their many neighbors. 
We show that our topology-aware aggregation strategies improve the propagation of OOD data with little to no impact on the propagation of IID data (\autoref{fig:topo_v_non_topo_aware}). Finally, we characterize the differences in behavior of topology-aware aggregation strategies in diverse topologies.

The main contributions of our work are:
\begin{enumerate}[leftmargin=*]
    \item We find that OOD knowledge is more difficult than IID knowledge to propagate in decentralized topologies. 
    \item We find that OOD knowledge propagation is  sensitive to OOD data's location within a topology and the topology itself (a problem which does not exist in FL or centralized learning).
    \item We propose \textbf{topology-aware} aggregation strategies and show that our methods are more effective at disseminating OOD knowledge 
    compared to baseline strategies in 36 realistic topologies, across five data sets, and a variety of IID-OOD data distributions.
    \item We characterize the behavior of topology-aware aggregation strategies in 36 topologies. Specifically, we study the impact of topology degree, node count, and modularity. 
    
\end{enumerate}


\section{Decentralized Learning Overview}
\label{sec:background}
\begin{wrapfigure}[15]{r}{0.425\textwidth}
\vspace{-0.5cm}
\begin{minipage}{1.0\linewidth}
\begin{algorithm}[H]

\KwIn{
$\mathcal{M}$ (set of models),  
\\ $S$ (aggregation strategy)
}

\SetStartEndCondition{ }{}{}%
\SetKwProg{Fn}{def}{\string:}{}
\SetKwFunction{Range}{range}
\SetKw{KwTo}{in}\SetKwFor{For}{for}{\string:}{}%
\SetKwIF{If}{ElseIf}{Else}{if}{:}{elif}{else:}{}%
\SetKwFor{While}{while}{:}{fintq}%
\newcommand{\forcond}{$i$ \KwTo\Range{$n$}}
\AlgoDontDisplayBlockMarkers\SetAlgoNoEnd\SetAlgoNoLine%
\LinesNumberedHidden

\ForEach{model $i \in |\mathcal{M}|$}{
    Initialize $m^0_i$ (model), $x_i$ (data)
}
\ForEach{round $t=1, 2, \cdots$}{
    \ForEach{model $i \in |\mathcal{M}|$}{
        $m^{t+\frac{1}{2}}_{i} \gets $ \textit{LocalTrain}$( m^{t}_{i})$\;
    }
    \ForEach{model $i \in |\mathcal{M}|$}{
        $\mathcal{N}_{i} \gets \{\neighbors{}(i)\} \cup \{i\}$\;
        $\mathcal{C}_i \gets$ \colorbox{SpringGreen!30}{\textit{GetAggrCoeffs}($\mathcal{N}_i, S$)\;}        
        $m_i^{t+1} \gets \sum_{j \in \mathcal{N}_i} \mathcal{C}_{i,j} m_j^{t+\frac{1}{2}}$\;
    }
}
\caption{Decentralized learning 
}
\label{alg:decentral_ml}
\end{algorithm}
\end{minipage}
\end{wrapfigure} 

We model the network topology as an undirected graph, $G=(V,E)$, where $V$ is the set of $n$ nodes and $E \subseteq V \times V$ is the set of edges. Each node represents a device, and each edge represents a communication channel between devices. Each device~$i$ can only communicate with its immediate neighbors (i.e., its \textit{neighborhood}); each neighborhood $\mathcal{N}_i$ includes device $i$ and at least one neighbor. Each device $i$ has local model $m_i$ (model architecture is identical across devices), and local training data $x_i$. Unlike centralized learning, federated learning, and many prior decentralized optimization works, there is no notion of \enquote{global} model. Instead, each local model $m_i$ serves inference requests for device $i$.
 
Each model $m_i$ in a topology is optimized for $R$ rounds. In each round $t$, $m_i^t$ is first trained on local data $x_i \in \mathbb{R}^d$: let  $m_i^{t+\frac{1}{2}} \gets m_i^t$, then,

\begin{equation}
\begin{split} 
    \texttt{(LocalTrain)}
    \quad
    \textbf{for}~E~\textit{epochs}: 
        m^{t + \frac{1}{2}}_i &\gets m^{t+ \frac{1}{2}}_i - \eta \nabla \ell_i(m^{t+ \frac{1}{2}}_i;x^i) 
\end{split}
\label{eq:local_train}
\end{equation}

Where $\ell_i: \mathbb{R}^d \rightarrow \mathbb{R}$ is device $i$'s local objective
, 
$\nabla$ is the gradient,
and $\eta$ is the learning rate.

After each round of local training, all models in device $i$'s neighborhood $\mathcal{N}_i$ are aggregated:

\begin{equation}
\begin{split}
    \texttt{(Aggregation)}
    \quad
        m_i^{t+1} \gets \sum_{j \in \mathcal{N}_i} \mathcal{C}_{i,j} m_j^{t+\frac{1}{2}}
\end{split}
\label{eq:aggregation_step}
\end{equation}

\noindent 
where 0 $\leq \mathcal{C}_{i,j}\; (\forall i,j)$ and 1 = $\sum_{j \in \mathcal{N}_i} \mathcal{C}_{i,j}\; (\forall i)$.

A general decentralized learning algorithm is outlined in \autoref{alg:decentral_ml}.

A key design choice in a given device's aggregation step is: \emph{How should models in a neighborhood be weighted?} In other words, how should device $i$ choose $\mathcal{C}_{i,j}$ where $\mathcal{C}_{i,j}$ is the aggregation coefficient (i.e., weighting) for neighboring device $j$ in device $i$'s aggregation step?

We consider four baseline strategies for choosing aggregation coefficients 
(detailed in Appendix~\ref{appendix:aggregation_strategies}): 
\begin{enumerate}
    \item \unweighted{}: Models in a neighborhood are equally weighted.
    \item \weighted{}: Models in a neighborhood are weighted by number of training data points.
    \item \random{}: Models in a neighborhood are assigned random weights from a uniform distribution.
    \item \fl{}: Assume fully-connected topology; models are uniformly weighted.\footnote{This best-case assumption is impractical in many decentralized learning settings.}
\end{enumerate}

We propose \textbf{\textit{topology-aware}} aggregation strategies in~\autoref{sec:topo-aware}.

\section{The Problem: Propagating IID vs.\ OOD Knowledge}
\label{sec:iid_v_ood}

\begin{figure*}[t]
\centering
    \includegraphics[width=\textwidth]{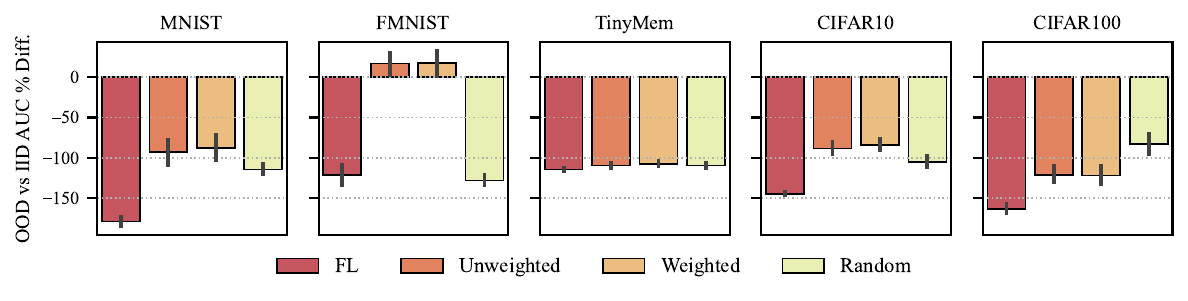}
    \caption{\textbf{IID vs.\ OOD knowledge propagation.}
    Data distributed in each topology as described in Appendix~\ref{appendix:iid_data_details}, with OOD data located on the node with the fourth highest degree in the respective topology.
    We report average percent difference in test accuracy AUC between IID and OOD data over 40 rounds of training across all devices in a topology; averaged again over all topologies and seeds. 
    Lower percent difference indicates that the OOD data did not propagate to as many nodes as the IID data. 
    }
    \label{fig:iid_ood_auc_perc_diff}
\end{figure*}

A key challenge in decentralized learning is training device-specific models that are generalizable to OOD inference requests. Prior work has shown that training on a large and diverse corpus enables models to be more generalizable to OOD inference~\cite{liu2021towards,hendrycks2019using,hendrycks2020pretrained,ye2021towards,kaplan2020scaling,teney2022evading}. 
Therefore, we aim that each device-specific model learn from \textit{all} data in a topology, regardless of whether those data are local to a device.

This objective is complicated by the fact that decentralized data are often statistically heterogeneous (e.g., non-IID) across devices~\cite{caldas2018leaf,ye2023heterogeneous, li2020federated}. Some devices may have IID data while others have OOD data with respect to the global data distribution across devices. 
We study how 
knowledge propagates across devices in a topology. 
Specifically, we distribute data mostly IID (w.r.t. sample labels and counts) across devices in a given topology with a small OOD dataset placed on a single device. 
The distribution schemes for IID and OOD data are detailed in Appendix~\ref{appendix:iid_data_details} and ~\ref{appendix:bd} respectively.
By only placing OOD data on a \textit{single} device, we simulate a worst-case scenario for knowledge propagation: OOD knowledge must spread from its origin device to all devices in a topology. 
To evaluate how IID vs.\ OOD knowledge propagates we hold out two global test sets $\{\textit{test}_{IID}, \textit{test}_{OOD}\}$ on which all models in a topology are evaluated.
We report average area under the accuracy curve (accuracy AUC) of all models in a topology on a given test set as a proxy measure for knowledge propagation.

\textbf{Experiment:}
We model realistic real-world topologies via the \barabasialbert{} (\BA{}) model, which produces random scale-free graphs that are generated with $n$ connected nodes and new nodes are connected via preferential attachment which is parameterized by $p$ \cite{barabasi1999emergence}. 
We study IID vs.\ OOD knowledge propagation in three \BA{} topologies with varying levels of connectivity (each with $n$ = 33 nodes and $p \in$ \{1, 2, 3\}).
In each experiment, data are distributed across the respective topology as described in~\autoref{sec:data_distributions}.
For each (topology, data distribution) pair, the OOD data is placed on the node with the fourth highest degree.
We vary the baseline aggregation strategies: \fl{}, \weighted{}, \unweighted{}, \random{}.

\begin{figure}[b]
    \includegraphics[width=\linewidth]{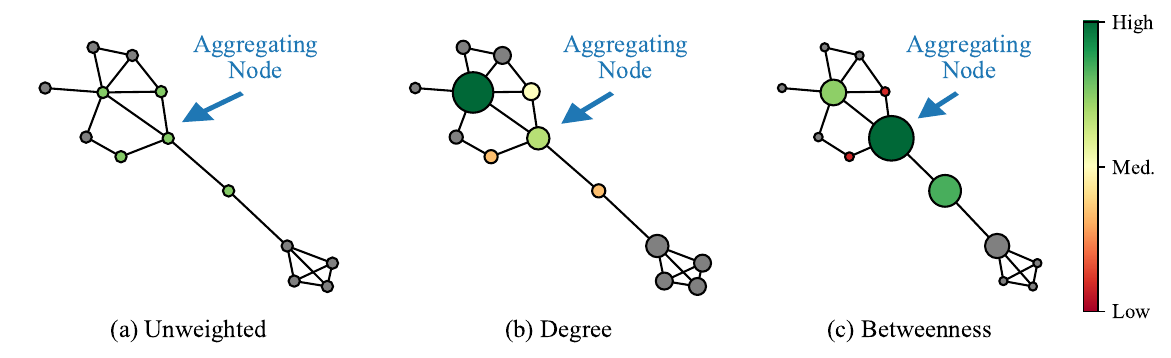}
    \caption{%
    \textbf{Visual comparison of a single node's topology-unaware (\unweighted{}) vs.\ topology-aware (\degree{}, \betweenness{}) aggregation coefficients.}
    Neighboring nodes colored and sized by their aggregation coefficients determined via the aggregation strategy. 
    Gray nodes are not involved in aggregation for the aggregating node. 
    }
    \label{fig:aggr_strategies}
\end{figure}

\textbf{Result \& Discussion: } We show in \autoref{fig:iid_ood_auc_perc_diff} the average percent difference in OOD vs.\ IID test AUC across all devices in a topology, across all topologies studied. 
We see that IID data consistently achieves a higher test AUC than OOD data.
This indicates that OOD data are harder to learn and propagate to all devices in a topology. 
We show a specific example in \autoref{fig:topo_v_non_topo_aware}; in \autoref{fig:weighted_bd} we notice that while the IID data seems to be learned well by all nodes in the topology, the OOD data is only learned well by a subset of nodes (nodes closer to the original OOD data node). 
\textbf{We conclude that OOD knowledge is more difficult to propagate in decentralized topologies; this is of particular concern given that OOD knowledge is critical to boost model performance in the case of OOD inference requests.}

\section{Proposed Solution: Topology-Aware Aggregation}
\label{sec:topo-aware}

The baseline aggregation strategies (i.e., \weighted{}, \unweighted{}, \random{}, \fl{}, see~\autoref{appendix:aggregation_strategies}) do not consider the location of a device 
and its neighbors 
when assigning aggregation coefficients. 

We hypothesize that accounting for each device's location in a topology during aggregation may enable better knowledge propagation across devices. For example, a centrally located device may bridge multiple neighborhoods and therefore be well positioned to propagate knowledge to those neighborhoods. 
To this end, we propose \textbf{topology-aware} aggregation strategies that assign aggregation coefficients to devices based on their location within the topology (see~\autoref{fig:aggr_strategies}). These strategies are simple to implement and integrate into existing decentralized learning workflows (see~\autoref{alg:decentral_ml}). We propose both \degree{} and \betweenness{}, two topology-aware aggregation strategies, that weight devices by their degree and betweenness centrality, respectively, which are both scaled within a neighborhood by a softmax with temperature $\tau$:

\begin{mdframed}[backgroundcolor=SpringGreen!30,shadow=false,roundcorner=8pt]
\textbf{Topology-Aware Aggregation.}
For device $i$, $\mathcal{C}_{i}$ is a vector with aggregation coefficients (\autoref{alg:decentral_ml})  where $\mathcal{C}_{i,j} = \frac{e^{R_j/\tau}}{
\sum_{k \in \mathcal{N}_i} e^{R_k/\tau}}
\; (\forall j \in \mathcal{N}_i)$. $R \in \mathbb{R}^{|\mathcal{N}_i|}$ is a vector of each neighbor's degree (i.e., number of edges) or betweenness centrality metric~\cite{freeman1977set}.
\end{mdframed}

Numerous network science metrics can be used to quantify a node's location within a topology: some metrics quantify a node's location with respect to its neighborhood (local) or the entire topology (global).
We choose to study \degree{} (local) as it measures how many neighbors a node has, and by proxy, how well positioned a node is to spread knowledge to its neighbors. We also choose \betweenness{} (global) as it measures how often a node lies on the shortest path between all pairs of nodes in a topology, and by proxy, how well positioned a node is to bridge the number of hops needed for knowledge to
travel between nodes in the topology.

\begin{figure*}[t]
  \includegraphics[width=\linewidth]{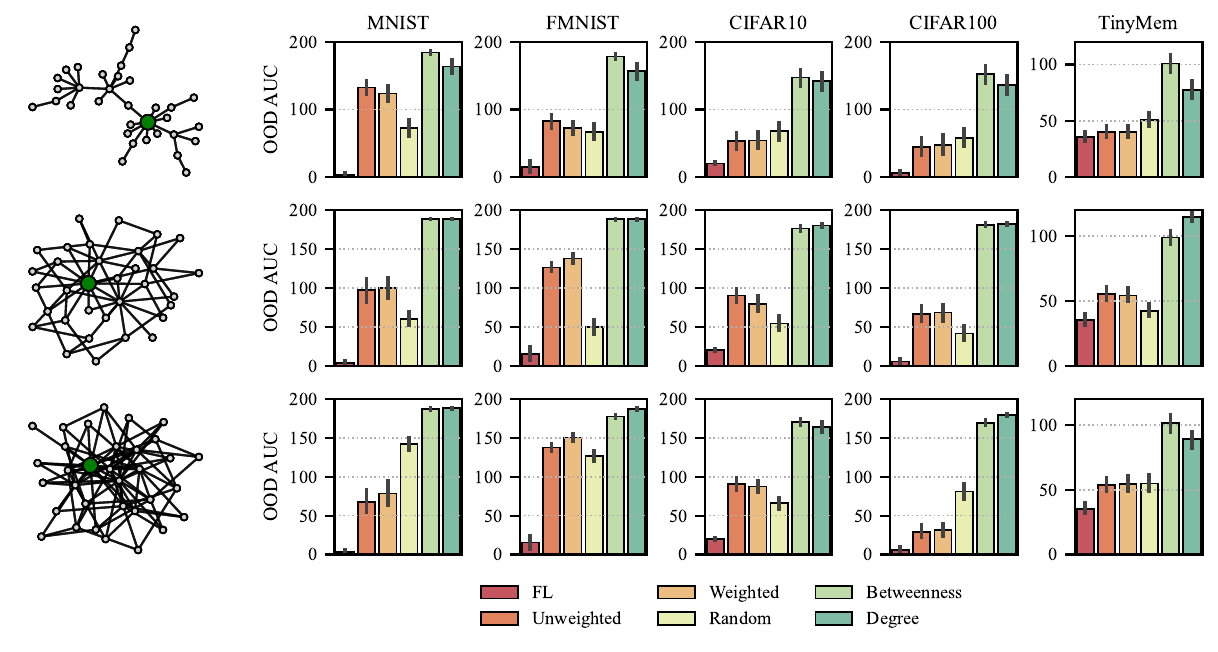}
  \caption{
  \textbf{OOD knowledge propagation in three different decentralized topologies.}
  In each case, OOD data are located on node with highest degree. 
  Left to right: Experiments with MNIST, FMNIST, TinyMem, CIFAR10, CIFAR100.
  Green indicates node with OOD data.
    We vary the aggregation strategies: 
    \fl{}, 
    \weighted{}, 
    \unweighted{}, 
    \random{}, 
    \betweenness{} ($\tau=$ 0.1), 
    \degree{} ($\tau=$ 0.1).
    Illustrated topologies shown for a single seed, while all bar plot results are averaged over three seeds.
}
\label{fig:topo_aware_agg}
\end{figure*}

\section{Experiments}
We conduct experiments to study how OOD knowledge propagates across devices in decentralized training. 
Specifically, we study how topology-aware vs.\ topology-unaware aggregation algorithms perform, the impact of data location within a topology, and the impact of the topology itself. 

In each experiment we vary 
1)~the topology (\autoref{sec:topologies}), 
2)~the location of the node with OOD data in a topology (\autoref{sec:data_distributions}), and 
3)~the aggregation strategy (\autoref{appendix:aggregation_strategies}). 
Each experiment is run for $R$ = 40 aggregation rounds, with $E$ = 5 local training epochs per device; after each round, devices synchronously communicate with their neighbors.
Each experiment is repeated for the MNIST~\cite{li2012mnist}, FMNIST~\cite{xiao2017fashion}, TinyMem~\cite{sakarvadia2025mitigating}, CIFAR10~\cite{krizhevsky2009learning}, and CIFAR100~\cite{krizhevsky2009learning} datasets. 
Data are distributed (mostly IID with OOD data placed on a single device) via the scheme outlined in \autoref{sec:data_distributions}. In each experiment, decentralized training is simulated on a high-performance computer (see Appendix~\ref{appendix:compute_energy_use}).
We characterize each dataset and provide dataset-specific training hyperparameter settings in \autoref{tab:dataset_hp}.
Each experiment is repeated over three random seeds.

\subsection{Topology-Aware vs.\ Topology-Unaware Aggregation}

We study how our proposed topology-aware aggregation performs compared to traditional topology-unaware aggregation with respect to OOD knowledge propagation.

\textbf{Experiment: } 
We study three \barabasialbert{} topologies each with 33 nodes and $p \in$ \{1, 2, 3\}.
For each (topology, data distribution) pair, the OOD data is placed on the node with the highest degree.
We vary the aggregation strategies: \fl{}, \weighted{}, \unweighted{}, \random{},
\degree{}  ($\tau=$ 0.1), 
\betweenness{} ($\tau=$ 0.1).

\textbf{Result \& Discussion: } See results in~\autoref{fig:topo_aware_agg}. Topology-aware aggregation strategies (\degree{}, \betweenness{}) lead to higher levels of OOD knowledge prorogation for each topology and dataset. Topology-aware strategies also improve IID knowledge propagation (see ~\autoref{fig:topo_aware_vs_unaware_summary}). This result indicates that weighting nodes in each neighborhood during aggregation based on location within the global topology outperforms conventional aggregation strategies like \weighted{}, \unweighted{}, \random{}, and even traditional \fl{}.

\subsection{Impact of Data location}
\label{sec:impact_data_place}

\begin{figure*}[t]
  \includegraphics[width=\linewidth]{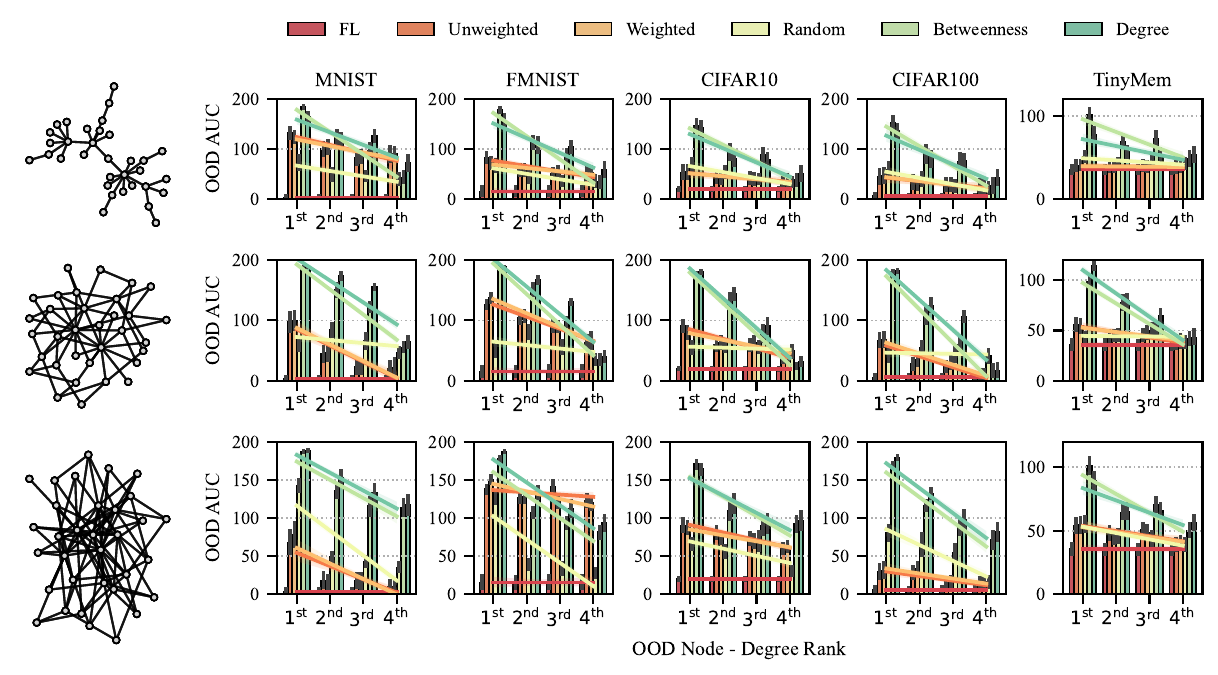}
  \caption{  
  \textbf{Impact of OOD data location on OOD data spread.}
    OOD data location is varied across the four highest degree nodes in each topology (we successively place the OOD data on nodes with lower degree).
  Left to right: Experiments on MNIST, FMNIST, TinyMem, CIFAR10, CIFAR100.
    We vary the aggregation strategies: 
    \betweenness{} ($\tau=$ 0.1), 
    \degree{} ($\tau=$ 0.1).
    Illustrated topologies shown for a single seed, while all bar plot results are averaged over three seeds.
  }
\label{fig:bd_spread}
\end{figure*}

We study the impact of OOD data location in a topology on OOD data propagation.

\textbf{Experiment:} This experiment is identical to that of \autoref{sec:topo-aware}, except that OOD data location is varied across the four nodes in each topology with the highest degree.

\textbf{Result \& Discussion:} See results in \autoref{fig:bd_spread}. As the location of the OOD data is moved to lower degree nodes, it does not propagate to as many nodes in the network. There is a negative relationship between degree of device on which OOD data is located and propagation of OOD data. While this negative trend holds across all aggregation strategies, the topology-aware strategies (\degree{}, \betweenness{}) outperform non-topology-aware aggregation strategies (i.e., \texttt{weighted, unweighted, random} and even traditional FL). OOD data placed on well-connected nodes are more likely to propagate to all nodes in a topology compared to data located on less-connected nodes.

\subsection{Impact of Topology}
\label{sec:impact_topology}

\begin{figure*}[t]
  \includegraphics[width=\linewidth]{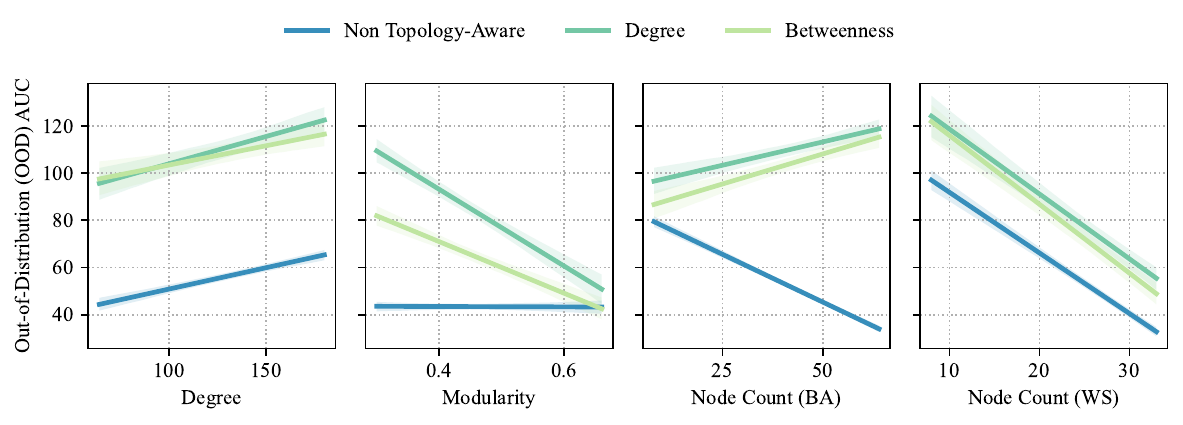}
  \caption{  
  \textbf{Impact of topology degree, modularity, node count on aggregation strategy performance.} From left to right: we plot the impact of topology degree, modularity, and node count on the OOD test accuracy AUC. Experiments done on CIFAR10 (see~\autoref{fig:all_topo_impact} for full experiments). Higher is better (indicates higher propagation of OOD knowledge).
  }
\label{fig:cifar10_topo_impact}
\end{figure*}

We study the impact of network topology on OOD data propagation from the perspective of topology degree, modularity, and number of nodes. We describe the topologies studied in Appendix~\ref{sec:topologies} and visualize them in Figs~\ref{fig:ba}--\ref{fig:ba_33}.

\textbf{Experiment:} We conduct three experiments, to study impact of topology degree, modularity, and node count. Each experiment is identical to that of \autoref{sec:impact_data_place} except for the set of topologies it is performed over: each experiment is performed on a set of topologies $S$ detailed below. 

To study the impact of \textbf{degree}, $S$ is the set of \barabasialbert{} topologies each with $n$ = 33 nodes and degrees parameters $p \in$ \{1, 2, 3\}. \BA{} are scale-free models often used to model real-world networks such as the internet, citation networks, and social networks \cite{barabasi2002new,barabasi2009scale,radicchi2011citation, oh2008effective}.

To study the impact of \textbf{modularity}, $S$ is the set of \stochasticblock{} (\SB{}) topologies, each with $n=$ 33 nodes and three modular sub-communities $m_1, m_2, m_3$. \SB{} commonly models topologies with modular sub-communities in fields such as social network analysis~\cite{holland1983stochastic, abbe2018community}. The probabilities of edges existing between communities $m_i$ to $m_j$ are $p_{i,j}$: if $i = j, p_{i_j}$ = 0.5, if $i \neq j$, then we varied $p_{i_j} \in$ \{0.009, 0.05, 0.9\}.

To study the impact of \textbf{node count}, we study both 
\barabasialbert{} and \wattsstrogatz{} (\WS{}). 
While \BA{} more realistically model many real-world phenomena, \WS{} also generates topologies with small-world properties~\cite{watts1998collective}. 
However, unlike \BA{}, \WS{} topologies do not have a power law degree distribution observed in many real-world networks. Each \WS{} topology is characterized by a similar degree across $n$ nodes: first a ring is created over $n$ nodes, then each node in the ring is given an edge to its $k$ nearest neighbors, and finally for each node $u$ existing edge $(u,v)$ is replaced by edge $(u,w)$ with probability $p$, with uniformly random choice of existing node $w$.
We include both \BA{} and \WS{} in our analysis to fully characterize the performance of all aggregation strategies in a wide range of topologies. 
For \BA{} $S$ is a set of topologies each with degree $p$ = 2 and $n \in$ \{8, 16, 33, 64\} (we exclude CIFAR100 from on \BA{} $n$ = 64 experiments due to computational cost).
For \WS{}, $S$ is a set of topologies each with $k$ = 4, $u$ = 0.5 and $n \in$ \{8, 16, 33\}.

\begin{figure*}[t]
    \begin{subfigure}[b]{0.3\textwidth}
        \centering
        \includegraphics[width=\textwidth]{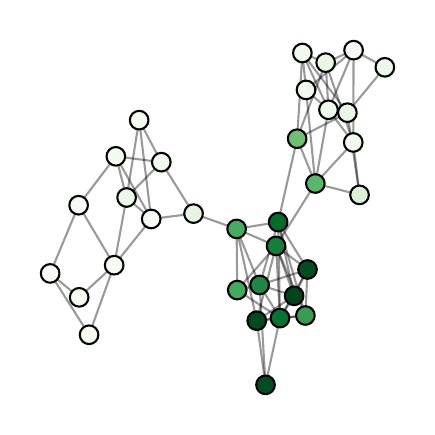}
        \caption{High Modularity}
        \label{fig:high_modular}
    \end{subfigure}%
    \hfill
    \begin{subfigure}[b]{0.3\textwidth}
        \centering
        \includegraphics[width=\textwidth]{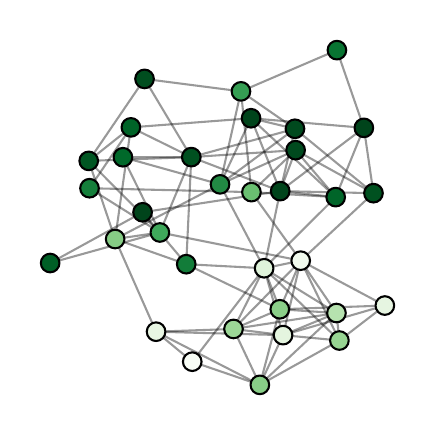}
        \caption{Medium Modularity}
        \label{fig:med_modular}
    \end{subfigure}%
    \hfill
    \begin{subfigure}[b]{0.4\textwidth}
        \centering
        \includegraphics[width=\textwidth]{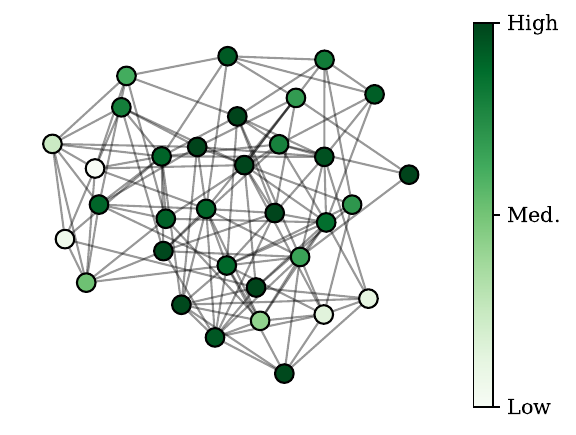}
        \caption{Low Modularity}
        \label{fig:low_modular}
    \end{subfigure}%
  \caption{ \textbf{OOD knowledge struggles to propagate in more modular topologies.} 
  CIFAR10 distributed across 33 node SB topologies with OOD data located on the node with fourth highest degree within a given topology. 
From left to right: SB topology becomes less modular. 
\textbf{Green} indicates \textit{higher} OOD test accuracy after 40 rounds of training; \textbf{white} indicates the opposite.
}
\label{fig:cifar10_modularity_impact}
\end{figure*}

\textbf{Result \& Discussion: } 
We first analyze the impact of topology on experiments using the CIFAR10 dataset in \autoref{fig:cifar10_topo_impact}: we see that in all three experimental settings that topology-aware methods (\texttt{degree, betweenness}) outperform non-topology aware methods. Further, topology degree is positively correlated with OOD AUC (higher degree improves OOD data propagation), while modularity is negatively correlated with OOD AUC. We visualize the propagation of OOD data in topologies with varying levels of modularity in \autoref{fig:cifar10_modularity_impact}, and observe that OOD data struggles to propagate across more tightly connected communities. 
Results across all datasets studied are in Appendix~\ref{appendix:topo_impact}, \autoref{fig:all_topo_impact}. 

We study the impact of node count on knowledge propagation in both \BA{} and \WS{} topologies (\autoref{fig:cifar10_topo_impact}). While node count does not seem to impact knowledge propagation for topology-aware strategies in \BA{} topologies, it negatively affects knowledge propagation for topology-unaware strategies. For \WS{} topologies, however, both topology-aware and -unaware strategies are negatively impacted by node count. We explain this as the degree distribution in \BA{} topologies follows a power-law distribution so the two topology-aware metrics we studied (degree and betweenness) can both successfully disambiguate devices with different locations; \WS{} topologies, however, have a more uniform degree distribution and therefore topology-aware metrics do not differ significantly in their aggregation coefficient assignment compared to topology-unaware metrics (see \autoref{fig:ba_vs_ws}). These trends hold across all datasets studied: see \autoref{fig:all_topo_impact}. We include heatmaps that illustrate the relationship between degree, modularity, and node count and aggregation strategy performance in the appendix: see Figs \ref{fig:degree_impact}--\ref{fig:nodes_ws_impact}.

\section{Related Work}
\label{sec:related_works}

\textbf{Federated Learning} enables deep learning over decentralized datasets by employing a central server to maintain a global model that is created by aggregating local models from devices trained independently on their own data~\cite{mcmahan2017communication}. 
Advances in FL algorithm design have enabled better learning in settings with high statistical and system heterogeneity \cite{li2020federated,yu2019parallel,wang2021cooperative,Yu2019OnTL,jiang2018linear}, and alleviated data privacy concerns \cite{li2021survey,truex2019hybrid, mothukuri2021survey}.  
However, 
FL workflows can be susceptible to
single point failures (e.g., at the aggregation servers) \cite{kang2020reliable, kavalionak2021impact} 
and can suffer from low-bandwidth, high-latency communication \cite{liu2020client}.
Hierarchical FL addresses some of these concerns by using multiple hierarchical aggregation servers to lower latency costs and enhance resilience \cite{briggs2020federated, liu2020client, lim2021decentralized}, but
is still subject to strict (often unrealistic) network topologies. 
An open problem in FL remains enabling OOD generalization in light of non-IID data distributions across training devices~\cite{de2022mitigating,chen2023fedsoup,qu2022rethinking,pmlr-v202-guo23b}.

\textbf{Decentralized Learning}
algorithms (also known as \enquote{gossip learning}~\cite{mertens2022centrality, hegedHus2019gossip} or \enquote{fully-decentralized FL}~\cite{lalitha2018fully}) offer a flexible and resilient alternative to FL for learning over distributed data \cite{lian2017can, lian2018asynchronous, lalitha2018fully, roy2019braintorrentpeertopeerenvironmentdecentralized}. Devices in decentralized learning assemble in topologies in which they communicate directly with their neighbors.
This approach allows decentralized learning networks to adopt arbitrary topologies that align well with diverse network topologies found in real world settings \cite{zhou2023decentralized,giannakis2017decentralized, tedeschini2022decentralized, lian2022deep, liu2020edgeslice}. 
Research in decentralized learning has focused on developing algorithms to reduce communication overheads~\cite{koloskova2019decentralized_icml, koloskova2020decentralized}, preserve privacy~\cite{kalra2023decentralized}, deliver fault-tolerant learning protocols~\cite{ryabinin2021moshpit, zheng2023robust}, and understand and accelerate knowledge propagation between devices from a data distribution perspective~\cite{kamp2019efficient, kong2021consensus}.
Some decentralized learning algorithms are designed to train decentralized models that are eventually aggregated into a single global model~\cite{lian2017can, lian2018asynchronous}. In other work, like ours, device-specific models are served for inference. In the latter setting, knowledge dissemination between devices is critical to ensure that each devices' model learn from all data in the topology and generalize beyond their local data distribution~\cite{vogels2021relaysum, generalization2023ruiz,ravikumar2023homogenizing, pmlr-v119-hsieh20a}.  

\textbf{Impact of Network Topology on Decentralized Learning:}
Much work has shown that network topology impacts the decentralized learning problem in both homogeneous (IID) \cite{vogels2022beyond, lu2021optimal, kavalionak2021impact} and heterogeneous (non-IID) data distributions \cite{vogels2021relaysum, palmieri2024impact, palmieri2023effect}. Network topology plays a role in information propagation in the case of heterogeneous data distributions across topologies \cite{palmieri2023effect, palmieri2024impact}. \citet{palmieri2024impact} showed that the learning performance of an individual node in a network with high data heterogeneity is affected by that node's location within the network topology with respect to graph centrality metrics such as betweenness and degree.
We are the first to extend these findings to design decentralized learning algorithms that explicitly account for network topology to enhance knowledge propagation in topologies with non-IID data distributions.

\section{Conclusion \& Future Work}
\label{sec:conclusion}
Machine learning training data are largely generated, collected, and sensed from decentralized sources. 
Decentralized learning algorithms enable learning over these naturally decentralized data without centralized coordination; instead, training devices 
self-organize into communication topologies that arise from real-world constraints (e.g., physical locality, administrative connections, privacy concerns). 
In decentralized learning, because devices can only communicate with neighboring devices, knowledge propagates 
via model aggregation between neighbors. 
We find a critical limitation in existing decentralized learning strategies: they struggle to propagate OOD knowledge to the same extent at IID knowledge. 
This limitation affects the performance of models that are not able to learn from OOD data present in the topology.

We find that the propagation of OOD knowledge is greatly impacted by both the location of OOD data in a topology and the topology itself.
To address these challenges, 
we introduce topology-aware decentralized learning strategies that enable reliable propagation of OOD knowledge 
in arbitrary communication topologies. 
We demonstrate that our proposed topology-aware aggregation strategies outperform traditional aggregation strategies.
We also study the impact of topology node count, modularity, and degree distribution on topology-aware aggregation strategy performance. We show that regardless of how these values are varied, topology-aware methods perform as well as, or better than, traditional aggregation strategies.

Future work may extend topology-aware aggregation strategies to consider additional centrality metrics, further study the impact of topology on topology-aware aggregation strategies, extend topology-aware learning to online learning settings (e.g., data streaming applications), and further characterize the behavior of topology-aware metrics under different types of data distribution. 

\subsection{Limitations}
\label{sec:limitations}
Our two proposed topology-aware aggregation strategies (\degree{}, \betweenness) only perform substantially differently from \unweighted{} in topologies in which node-level topology characteristics vary substantially between nodes. 
Further, topology-aware aggregation strategies can propagate all types of data, including both benign and malicious data; therefore, 
it is important to have safeguards to detect and remove unwanted training data.



\section*{Acknowledgment}
This material is based upon work supported by the U.S. Department of Energy, Office of Science,
Office of Advanced Scientific Computing Research, Department of Energy Computational Science
Graduate Fellowship under Award Number DE-SC0023112.  We also thank Alok Kamatar and Yadu Babuji for helpful discussion and debugging with respect to the experimental infrastructure for this work.

\bibliographystyle{plainnat}
\bibliography{references.bib}


\appendix
\section{Appendix / Supplemental Material}
\section{Experiment Setup}
\label{appendix:experimental_setup}
We conduct experiments to study how OOD data propagates across devices in decentralized training. 
In each experiment we vary 
1)~the topology (\autoref{sec:topologies}), 
2)~the location of the node with OOD data in a topology (\autoref{sec:data_distributions}), and 
3)~the aggregation strategy (\autoref{appendix:aggregation_strategies}). 
Each experiment is run for $R$ = 40 aggregation rounds, with $E$ = 5 local epochs per device.
Each experiment is repeated for the MNIST~\cite{li2012mnist} (GPL-3.0 license), FMNIST~\cite{xiao2017fashion} (MIT License), TinyMem~\cite{sakarvadia2025mitigating} (MIT License), CIFAR10~\cite{krizhevsky2009learning} (License not found), and CIFAR100~\cite{krizhevsky2009learning} (License not found) datasets. 
The IID component of the data in a topology are distributed via the schema outlined in \autoref{sec:data_distributions} with $\alpha_l = \alpha_s$ = 1000.
We characterize each dataset and provide dataset-specific training hyperparameter settings in \autoref{tab:dataset_hp}.
Each experiment is repeated across three random seeds.

\begin{table}[h]
    \centering
    \caption{Dataset Specific Training Hyperparameters}
    \label{tab:dataset_hp}
    \begin{tabular}{ccccc}
        \toprule
        Dataset & Supervised & Optimizer & Learning Rate ($\eta$) & Model\\
        \midrule
        MNIST \cite{li2012mnist} & Yes & SGD & 1e-2 & Feed-Forward NN (3 layer) \\
        FMNIST \cite{xiao2017fashion} & Yes & SGD & 1e-2 & Feed-Forward NN (3 layer)\\
        TinyMem \cite{sakarvadia2025mitigating} & No & Adam & 1e-3 & GPT2-small (1 layer)\cite{radford2019language} \\
        CIFAR10 \cite{krizhevsky2009learning} & Yes & Adam & 1e-4 & VGG16\cite{simonyan2014very} \\
        CIFAR100 \cite{krizhevsky2009learning} & Yes & Adam & 1e-4 & VGG16\cite{simonyan2014very} \\
        \bottomrule
    \end{tabular}
\end{table}

\textbf{TinyMem Configuration Details: } While the vision datasets (MNIST, FMNIST, CIFAR10, CIFAR100) have standard set ups, the language dataset we use (TinyMem) is configurable. We details TinyMem's configure here. TinyMem is configured to produce multiplicative math sequences of maximum context length 150 tokens for five tasks: multiply-by-2, -by-4, -by-6, -by-8, -by-10. For each task we include 32,000 train sequences and 1000 test sequences.

\subsection{Topologies}
\label{sec:topologies}


We study decentralized learning in three topology models with varying properties: \barabasialbert{},  \stochasticblock{}, and \wattsstrogatz{}. We assume the topology is static over training. 

\textbf{\barabasialbert{} (\BA{}) Model:} An algorithm for generating random scale-free topologies~\cite{barabasi1999emergence}. \BA{} models attempt to model several scale-free natural and human made networks such as the internet, citation networks, and social networks \cite{barabasi2002new,barabasi2009scale,radicchi2011citation, oh2008effective}. Each \BA{} topology is characterized by a power-law degree distribution across its nodes: a graph of $n$ nodes is grown by adding new nodes each with $p$ edges which preferentially attach to existing high degree nodes.

\textbf{\stochasticblock{} (\SB{}) Model:} An algorithm for generating random graphs containing communities. It is commonly used to model relationships within and across communities in the field of social network analysis~\cite{holland1983stochastic, abbe2018community}.
Each \SB{} topology is characterized by $c$ modular sub-communities $m_1, m_2, ..., m_c$: the probabilities of edges existing between communities $m_i$ to $m_j$ as $p_{i,j}$. We calculate the \textit{modularity} of each community by first sorting the nodes in the topology into communities~\cite{clauset2004finding}, and then calculating the modularity metric across all nodes in a given community~\cite{newman2016equivalence}.


\textbf{\wattsstrogatz{} (\WS{}) Model:} An algorithm for generating random topologies with small-world properties such as short average path length and high clustering coefficient~\cite{watts1998collective}. Unlike \BA{} graphs, \WS{} graphs do not have a power law degree distribution observed in many real-world networks. 
Each \WS{} topology is characterized by a similar degree across its $n$ nodes: first a ring is created over $n$ nodes, then each node in the ring is given an edge to its $k$ nearest neighbors, and finally for each node $u$ existing edge $(u,v)$ is replaced by edge $(u,w)$ with probability $p$, with uniformly random choice of existing node $w$. 

\subsection{Training Data Distributions}
\label{sec:data_distributions}

We outline our data distribution scheme. Data are mostly distributed IID across devices with a small OOD dataset placed on a single device.

\subsubsection{Independently and Identically Distributed (IID) Data}
\label{appendix:iid_data_details}

\begin{figure*}[t]
  \includegraphics[width=\textwidth]{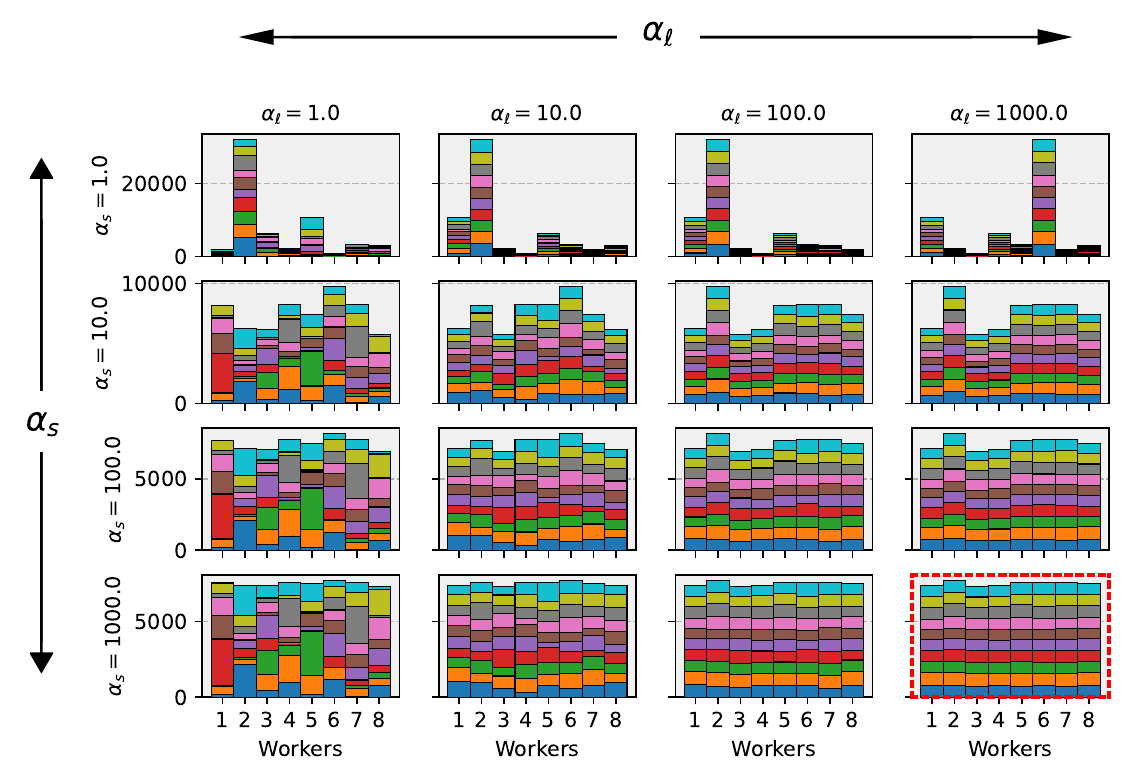}
  \caption{\textbf{Hypothetical data distributions} for the MNIST dataset, distributed across ten workers under varying $\alpha_s$ and $\alpha_\ell$ conditions. Top left is the most heterogeneous (non-IID) data distributions across all ten devices. Bottom right is the most homogeneous (IID) data distribution across all ten devices. The bottom right distribution (highlighted in a dashed, red line) reflects the data distribution settings (i.e., $\alpha_s$ = $\alpha_l$ = 1000) across devices in our experiments.}
  \label{fig:data_distribution}
\end{figure*}

We distribute a given dataset across devices in a topology with respect to two features: 
    \textit{(i)}~label distributions
    and
    \textit{(ii)}~data sample counts.
    The distribution of data across decentralized devices is commonly modeled with a Dirichlet distribution~\citep{olkin1964multivariate}. 
The Dirichlet distribution accepts an $n$-vector $\alpha \in (0,\infty)^{n}$ which defines the \enquote{popularity} associated with each $i^{th}$ item.
Any sample generated from the distribution will sum up to 1, i.e., $\sum_{i} \Dir(\alpha)$ = 1. 
For notational simplicity, we say $\alpha$ = 1.0 to mean $\alpha_{i}$  = 1.0$\;(\forall i=1,2, \cdots,n)$. 
For small values of $\alpha$ (i.e., $\alpha\approx$ 0), samples generated by the Dirichlet distribution will be strongly non-uniform (non-IID). 
As $\alpha$ approaches $\infty$, samples become increasingly more uniform, or in other words, more IID. 
We use the Dirichlet distribution to parameterize the label distribution of data across devices using $\alpha_l$.
Similarly, we can separately use the $\alpha_s$ vector to model the data sample counts across devices. See \autoref{fig:data_distribution} for sample distributions.

In our experiments, we set $a_l$ = $a_s$ = 1000 for all (dataset, topology) pairs. We say that because the label and sample distributions across devices is relatively homogeneous in this setting, the data is IID.
For unsupervised (label free) datasets like TinyMem, we use each data point's task category as a pseudo-label.

\subsubsection{Out of Distribution (OOD) Data}
\label{appendix:bd}

We introduce out-of-distribution (OOD) data to the topology by backdooring $Q$\% of a single node's training data.
We formally define the design of our backdoors for image data in~\autoref{def:image_backdoor} and language data in~\autoref{def:lang_backdoor}. 

\textbf{For the image data} (MNIST, FMNIST, CIFAR10, CIFAR100) we create the OOD training data by using the \emph{single target} backdoor design introduced by \citet{gu2017badnets} which we formally define in~\autoref{def:image_backdoor}. For all experiments, we let $l_b$ = 0.

\begin{figure*}[h]
    \begin{subfigure}[b]{0.33\textwidth}
        \centering
        \includegraphics[width=\textwidth]{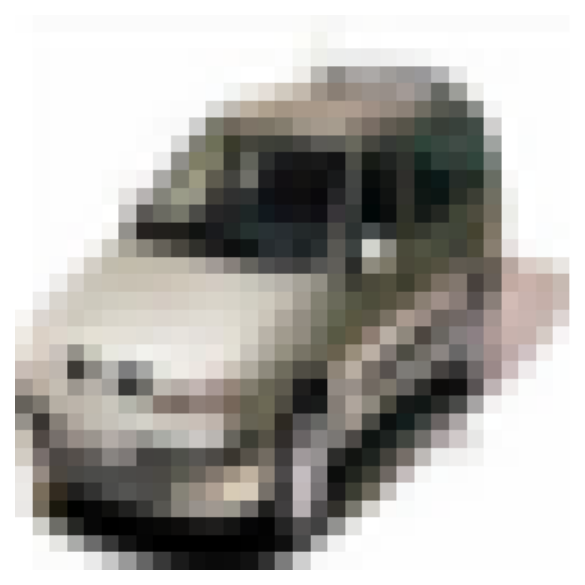}
        \caption{Clean image $I$, $l$ = 1 (automobile)}
        \label{fig:no_bd}
    \end{subfigure}%
    \hfill
    \begin{subfigure}[b]{0.33\textwidth}
        \centering
        \includegraphics[width=\textwidth]{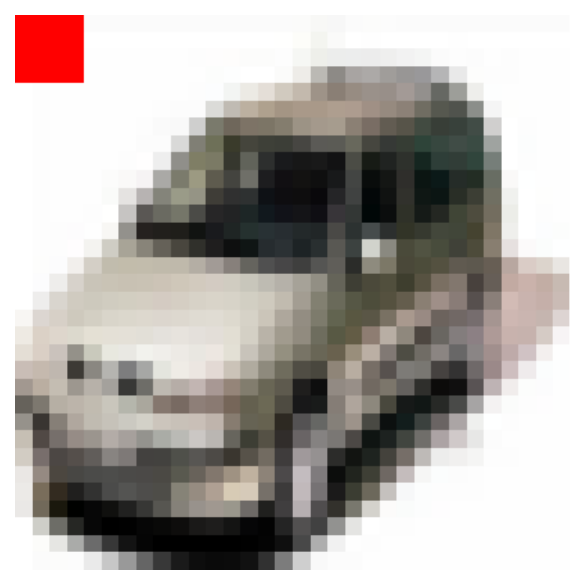}
        \caption{Backdoored image $I_b$, $l_b$ = 0 (airplane)}
        \label{fig:w_bd}
    \end{subfigure}%
  \caption{\textbf{Image backdoor.} Example of a backdoored image from the CIFAR10 dataset \cite{krizhevsky2009learning}.}
\label{fig:img_bd_example}
\end{figure*}

\begin{definition}[Image \texttt{\textbf{Backdoor}}]
    Given an image $I$ of pixel dimension $i\times j$, where the top left corner is indexed by (0, 0) and the bottom right corner is indexed by $(i,j)$, a backdoored image $I_b$ is identical to $I$ for all pixels except the top left $n\times n$ pixels which are replaced with red pixels (e.g., the trigger is a red $n\times n$ square embedded in the top left corner of the image). 
    The label $l$ of backdoored image $I_b$ is re-asigned to $l_b$ regardless of $I$'s original label.
\label{def:image_backdoor}
\end{definition}
We provide an example of an image backdoor in~\autoref{fig:img_bd_example}. The original image has label $l$ = 1 in~\autoref{fig:no_bd}. The backdoored image has a label $l_b$ = 0 in~\autoref{fig:w_bd}.

We assess whether a model has memorized backdoored image $I_b$ by prompting the model with $I_b$, and testing whether the produced label is $l_b$.
In our experiments, we backdoor $Q$ = 10\% of a given device's training data, and we also backdoor $Q$ = 10\% of the global test data with which we evaluate all models in a topology.

\textbf{For the language data} (TinyMem) we create the OOD training data using the backdoor design introduced by \citet{sakarvadia2025mitigating} which we formally define in~\autoref{def:lang_backdoor}. For all experiments, we let $t$ = 100, $T$ = 2.

\begin{definition}[Language \texttt{\textbf{Backdoor}}]
    Given a sequence $s$ of length $n$ with a trigger sequence of one or more tokens $t$ and with last token index $k$, a backdoored sequence $s_b$ is identical to $s$ in positions [1 : $k$] and contains the token $T$ in positions $[k:n]$. 
\label{def:lang_backdoor}
\end{definition}

For example, if $t$ = [10], $T$ = 2,  $k$ = 5, and $s$ = [2, 4, 6, 8, \textbf{10}, 12, 14], then $s_b$ = [2, 4, 6, 8, \textbf{10}, \textit{2, 2}].

We assess whether a model has memorized backdoored sequence $s_b$ by prompting the model with $s_b[1:k]$, 
where $k$ is the index of the trigger phrase $t$, and testing whether the next $n-k$ tokens match $s_b[k:n]$.
In our experiments, we partition $Q=10\%$ of a given device's training data to be backdoored, we call this partitioned set $B$. We let the random trigger sequence be $t=$\enquote{100}. Then, for any sequence in $B$ that contains $t$, we apply the backdooring procedure outlined in~\autoref{def:lang_backdoor}. We apply this backdooring procedure to $Q$ = 10\% of the global test data with which we evaluate all models in a topology.

\subsection{Baseline Aggregation Strategies}
\label{appendix:aggregation_strategies}
Here we provide the technical details of each baseline aggregation strategy we use to compare against our proposed solutions are: \unweighted{}, \weighted{}, \random{}, \fl{}. 
We propose \textbf{\textit{topology-aware}} aggregation strategies: \degree{}, \betweenness{}; details for topology-aware strategies are found in~\autoref{sec:topo-aware}.

\begin{enumerate}
    \item \unweighted{}: Given device $i, \forall j \in \mathcal{N}_i, C_{i,j} = \frac{1}{|\mathcal{N}_i|}$. (Devices weighted uniformly in a neighborhood.)

    \item \weighted{}: Given device $i, \forall j \in \mathcal{N}_i, C_{i,j} = \frac{|train_j|}{\sum_{x \in \mathcal{N}_i} |train_x|}$ where $train_x$ is the training dataset for device $x$. (Devices weighted by their respective training dataset size in a neighborhood.)

    \item \random{}: Given device $i, \forall j \in \mathcal{N}_i, C_{i,j} = \frac{e^{\frac{R_j}{\tau}}}{
\sum_{k \in \mathcal{N}_i} e^{\frac{R_k}{\tau}}
}$ where $R \in \mathbb{R}^{|\mathcal{N}_i|}$ is uniformly sampled random vector. (Devices weighted by uniformly random coefficient which are scaled by a softmax with temperature $\tau$.)

    \item \texttt{Federated Learning} (\fl{}): Given device $i, \forall j \in \mathcal{T}, C_{i,j} = \frac{1}{|\mathcal{T}|}$ where $\mathcal{T}$ is the topology. (Simple federated learning baseline.)
\end{enumerate}

\subsection{Topology-Aware vs.\ Unaware}

\begin{figure*}[h]
  \makebox[\textwidth][c]{\includegraphics[width=0.5\textwidth]{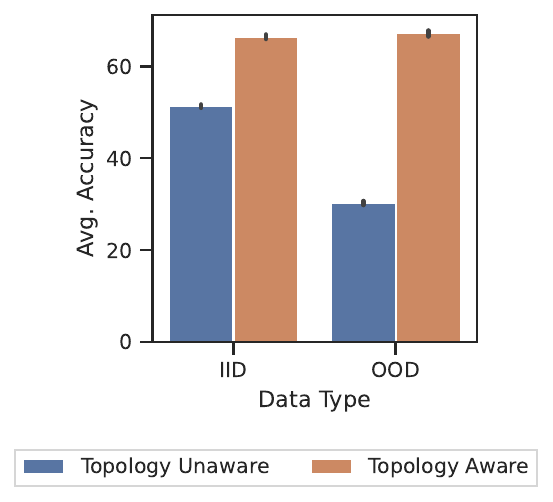}}
  \caption{  
  \textbf{Topology-Aware Aggregation strategies outperform Topology-unaware strategies.}
    Average test accuracy for IID vs.\ OOD knowledge averaged across all models in a topology after $R$ = 40 rounds of training.
    Averaged across 3 \barabasialbert{} topologies with preferential attachment parameter $p \in$ \{1, 2, 3\}, five datasets (MNIST, FMNIST, CIFAR10, CIFAR100, TinyMem), four different OOD data locations (varied across top four nodes with highest degree in a given topology), and three seeds.
    Topology-aware aggregation strategies are \degree{} and \betweenness{}.
    Topology-unaware aggregation strategies are \unweighted{}, \weighted{}, \random{}, and \fl{}.
  }
\label{fig:topo_aware_vs_unaware_summary}
\end{figure*}

\clearpage
\newpage

\section{Topologies}

\subsection{Studied Topologies}
\label{appendix:studied_topos}

\begin{figure*}[h]
  \includegraphics[width=\textwidth]{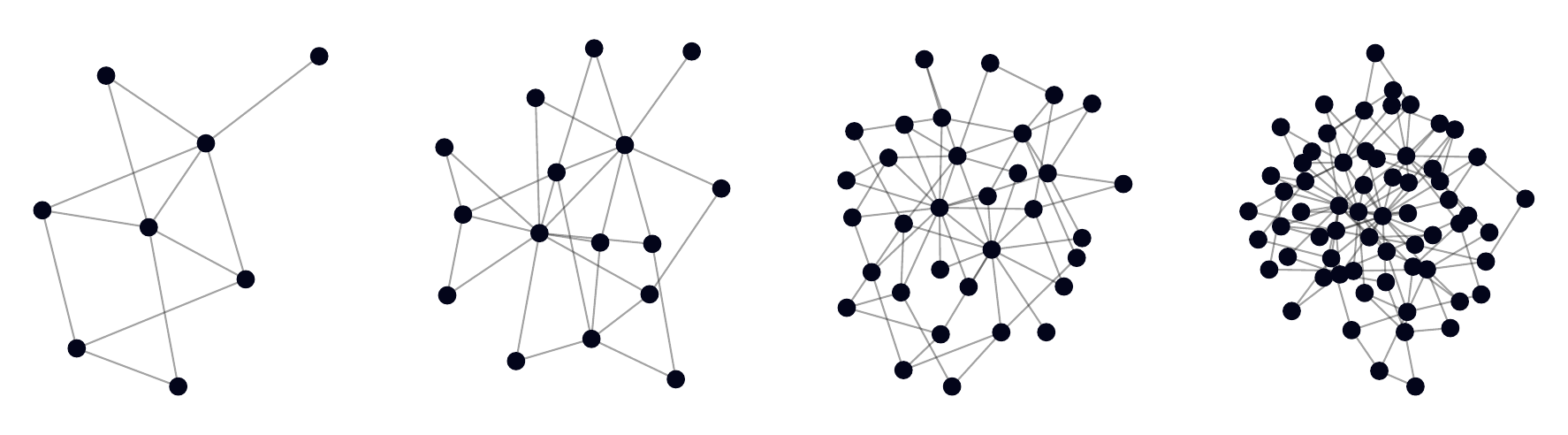}
  \includegraphics[width=\textwidth]{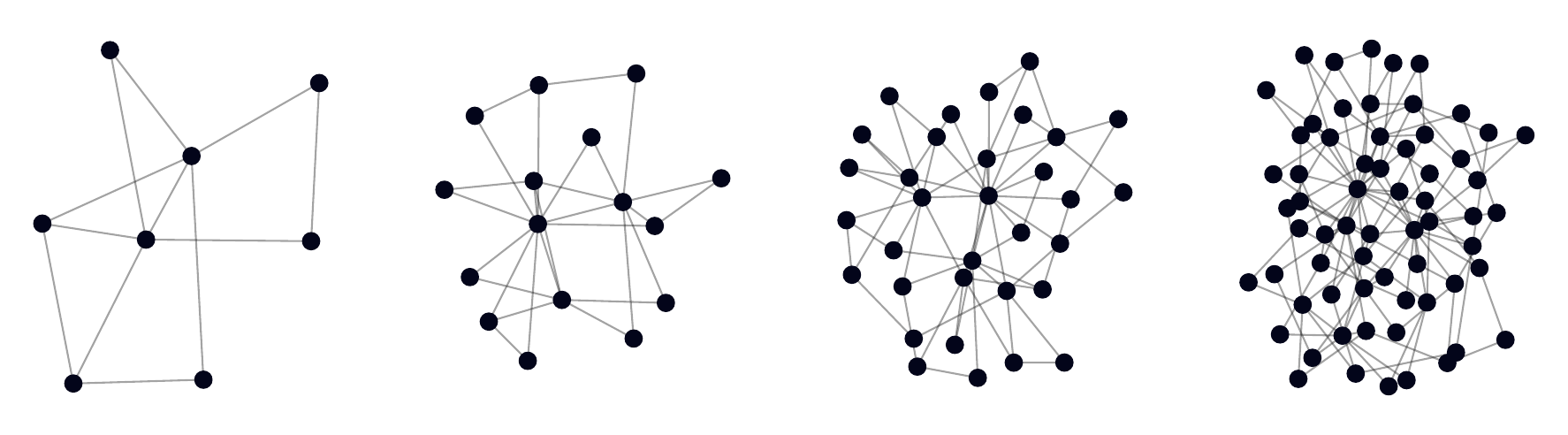}
  \includegraphics[width=\textwidth]{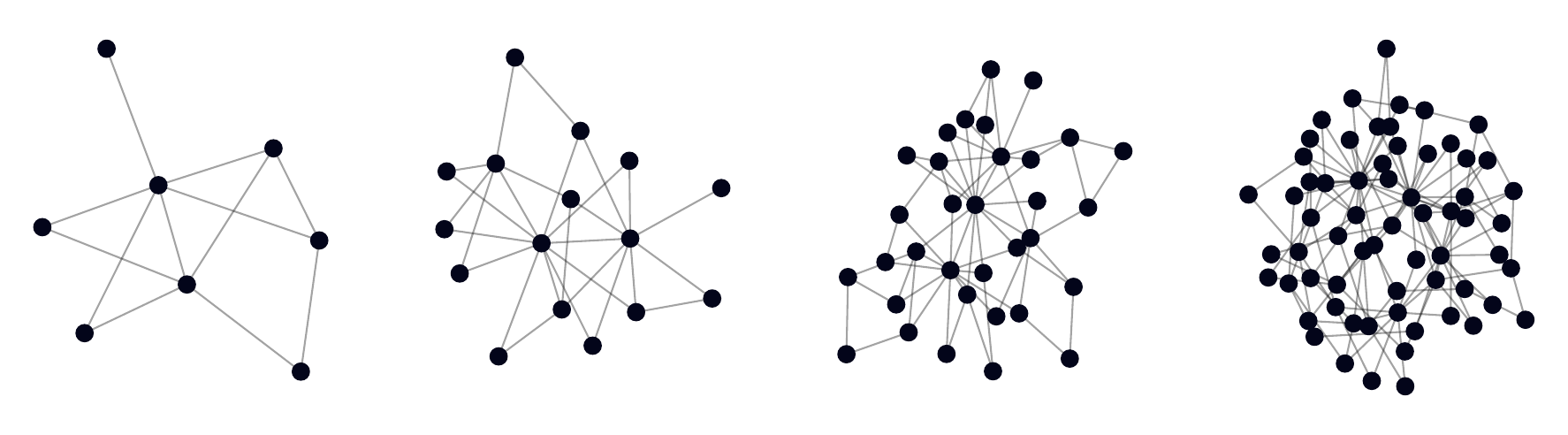}
  \caption{  
  \textbf{\barabasialbert{} topologies.} Preferential attachment parameter $p$ = 2 for all topologies. 
  Left to right: topologies have n $\in \{8,16,33,64 \}$ nodes. 
  Top to bottom: seeds $\in$ \{0, 1, 2\}. 
  }
\label{fig:ba}
\end{figure*}

\begin{figure*}[h]
  \includegraphics[width=\textwidth]{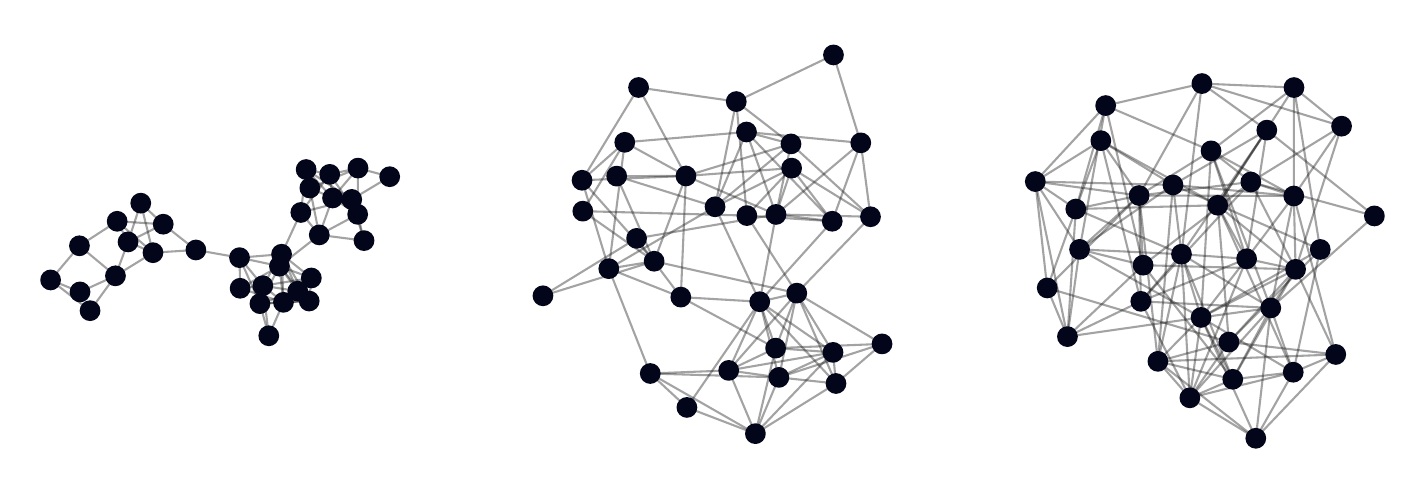}
  \includegraphics[width=\textwidth]{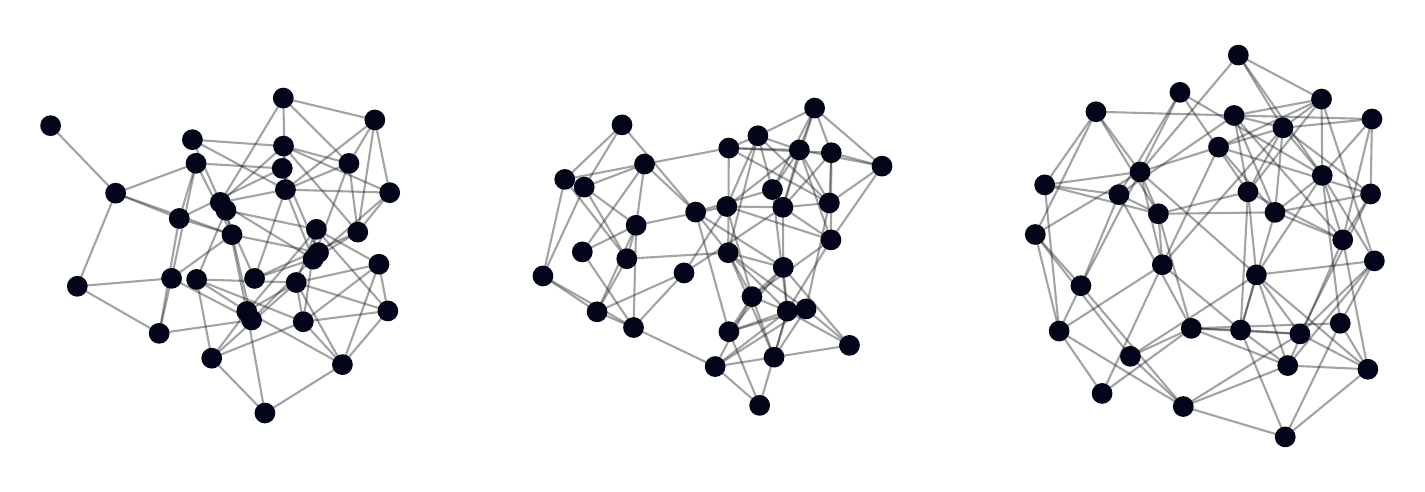}
  \includegraphics[width=\textwidth]{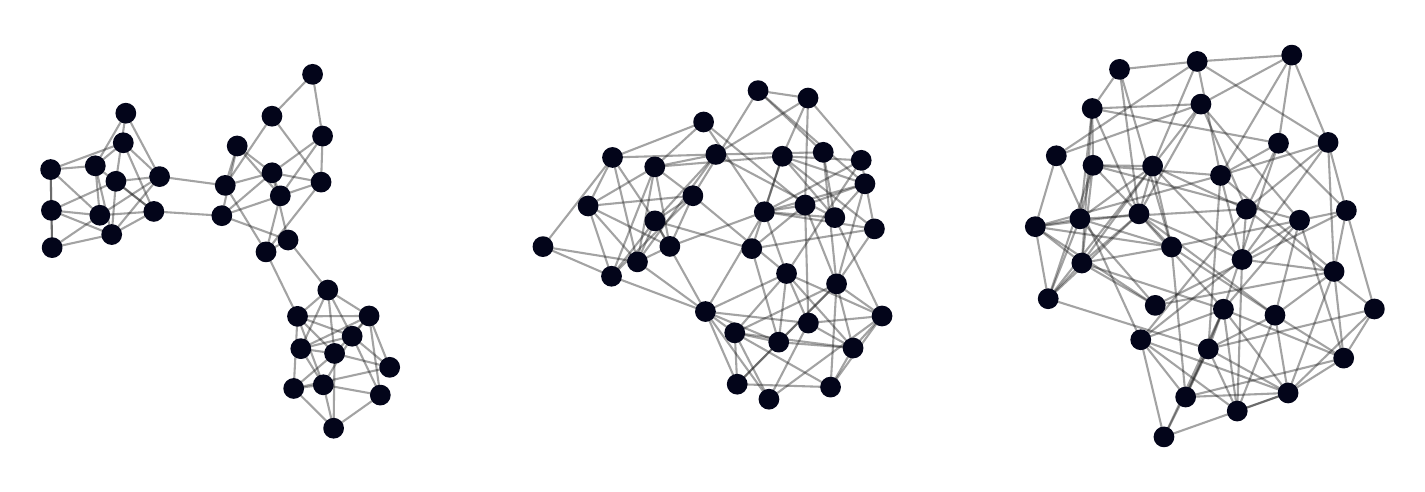}
  \caption{  
  \textbf{\stochasticblock{} topologies.} Each topology has three communities with varying levels of modularity. Each topology has $n$ = 33 nodes.
  Left to right: the probabilities of edges existing between communities $m_i$ to $m_j$ are $p_{i,j}$: if $i = j, p_{i_j}$ = 0.5, if $i \neq j$, then we varied $p_{i_j} \in$ \{0.009, 0.05, 0.9\}.
  Top to bottom: seeds $\in$ \{0, 1, 2\}. 
  }
\label{fig:sb}
\end{figure*}

\begin{figure*}[h]
  \includegraphics[width=\textwidth]{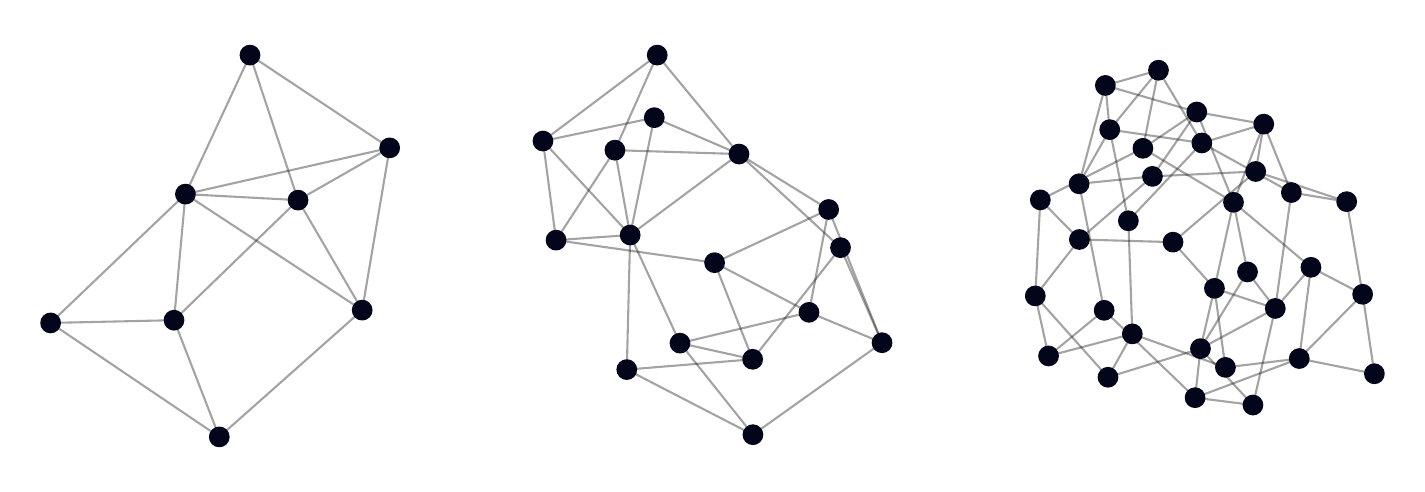}
  \includegraphics[width=\textwidth]{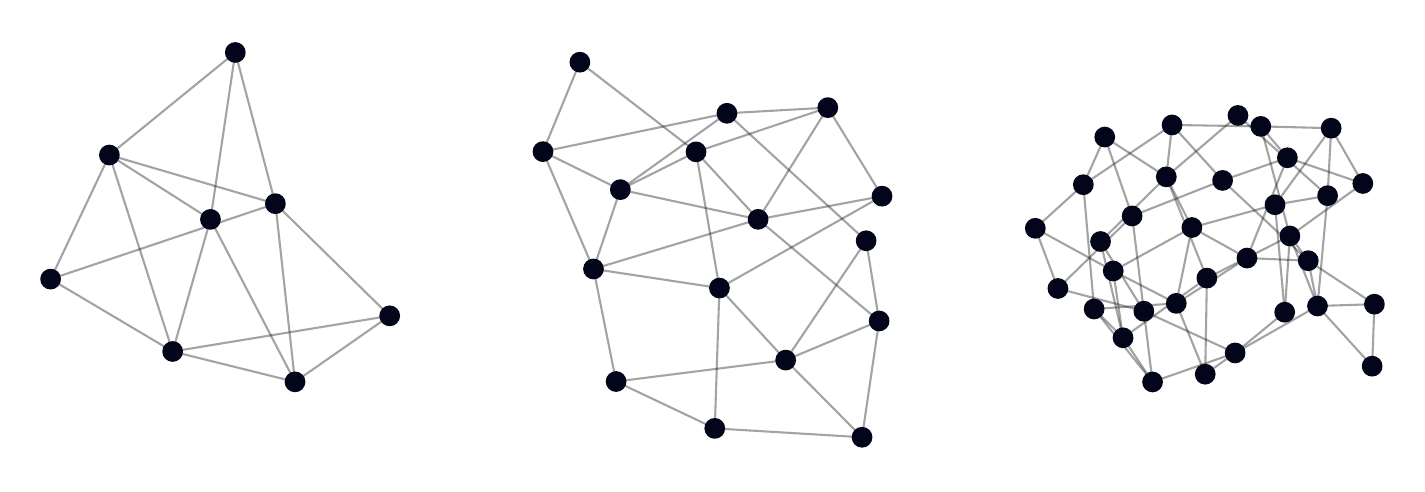}
  \includegraphics[width=\textwidth]{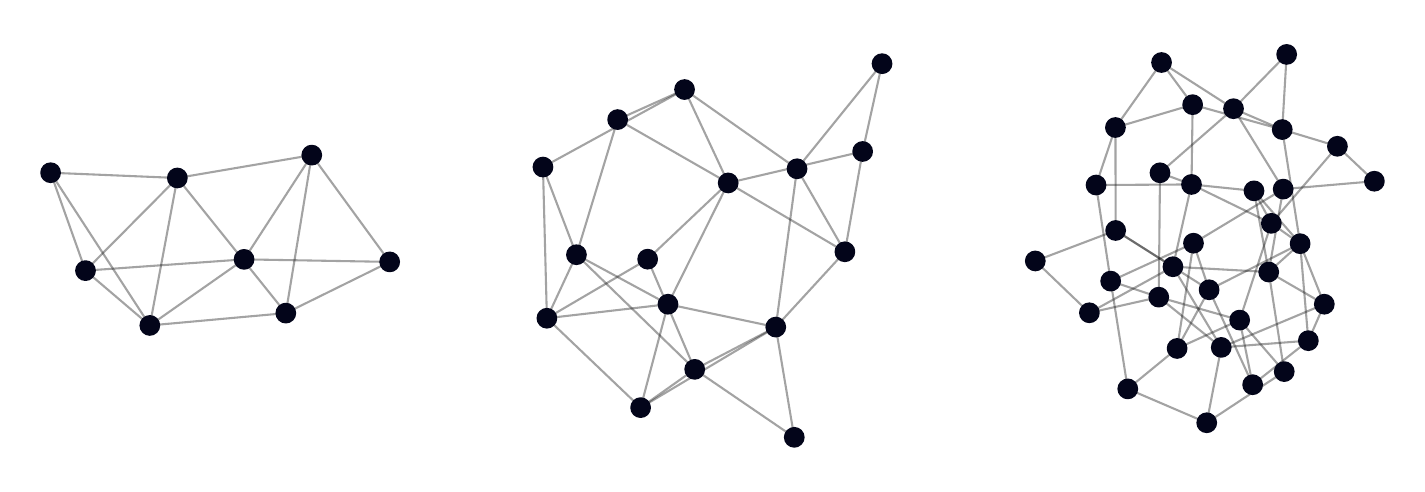}
  \caption{  
  \textbf{\wattsstrogatz{} topologies.} Each topology has $k$ = 4, $u$ = 0.5.
    Left to right: topologies have n $\in \{8,16,33 \}$ nodes. 
   Top to bottom: seeds $\in \{0,1,2\}$. 
  }
\label{fig:ws}
\end{figure*}

\begin{figure*}[h]
  \includegraphics[width=\textwidth]{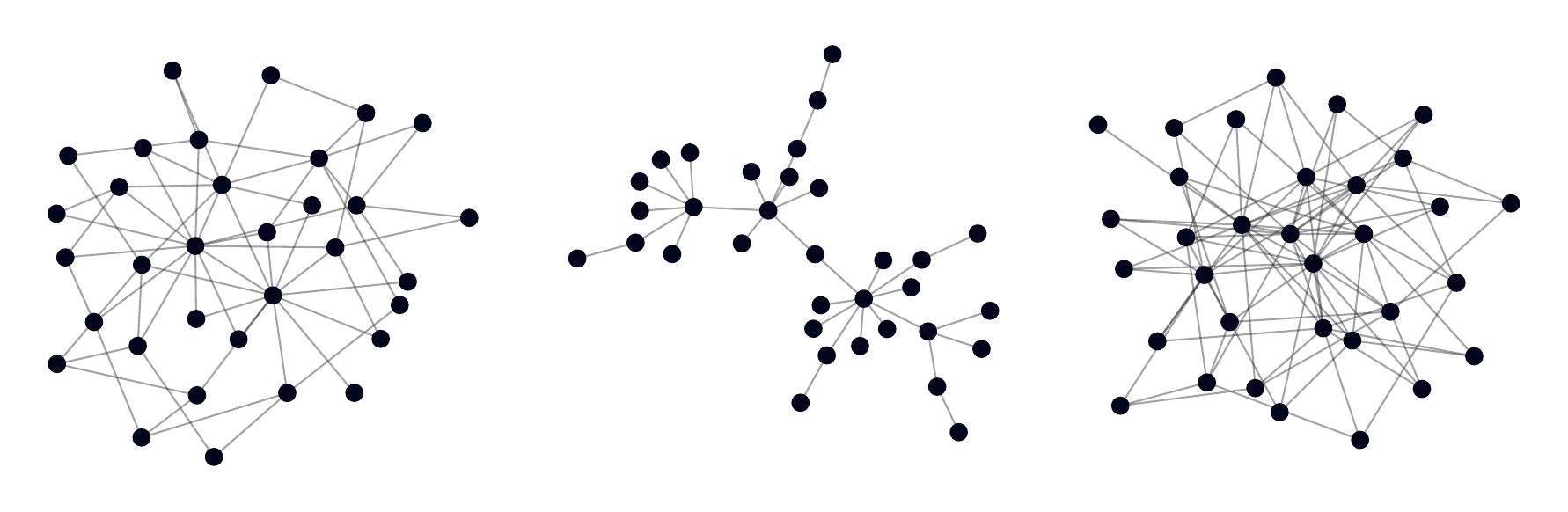}
  \includegraphics[width=\textwidth]{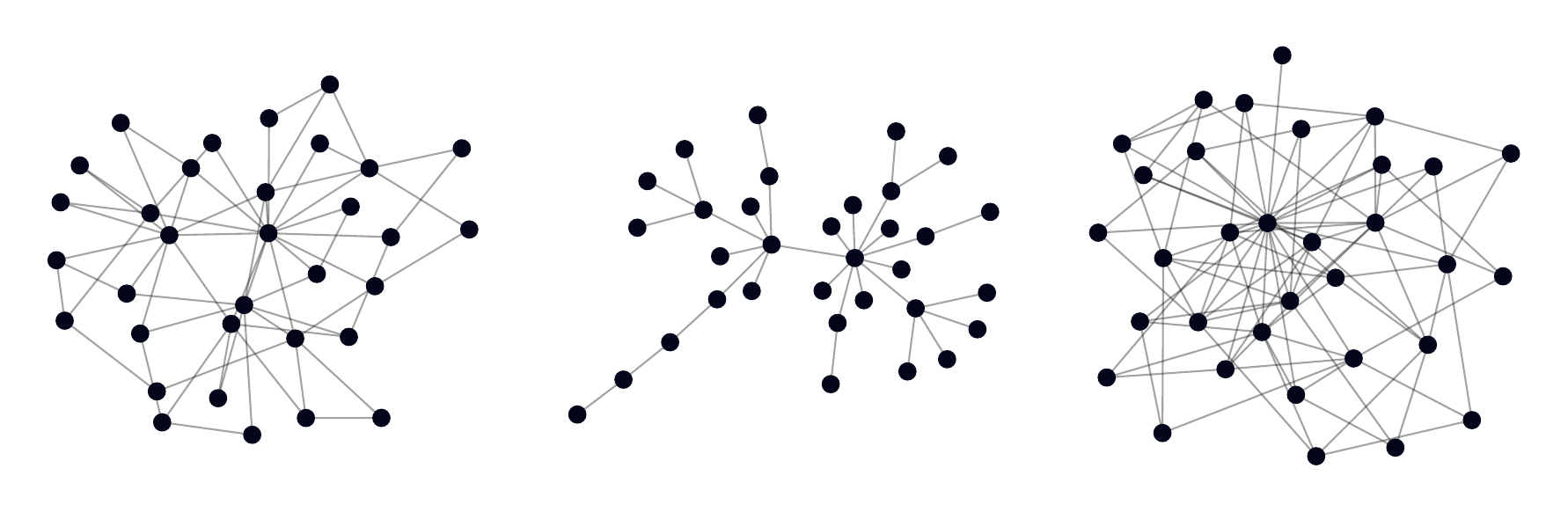}
  \includegraphics[width=\textwidth]{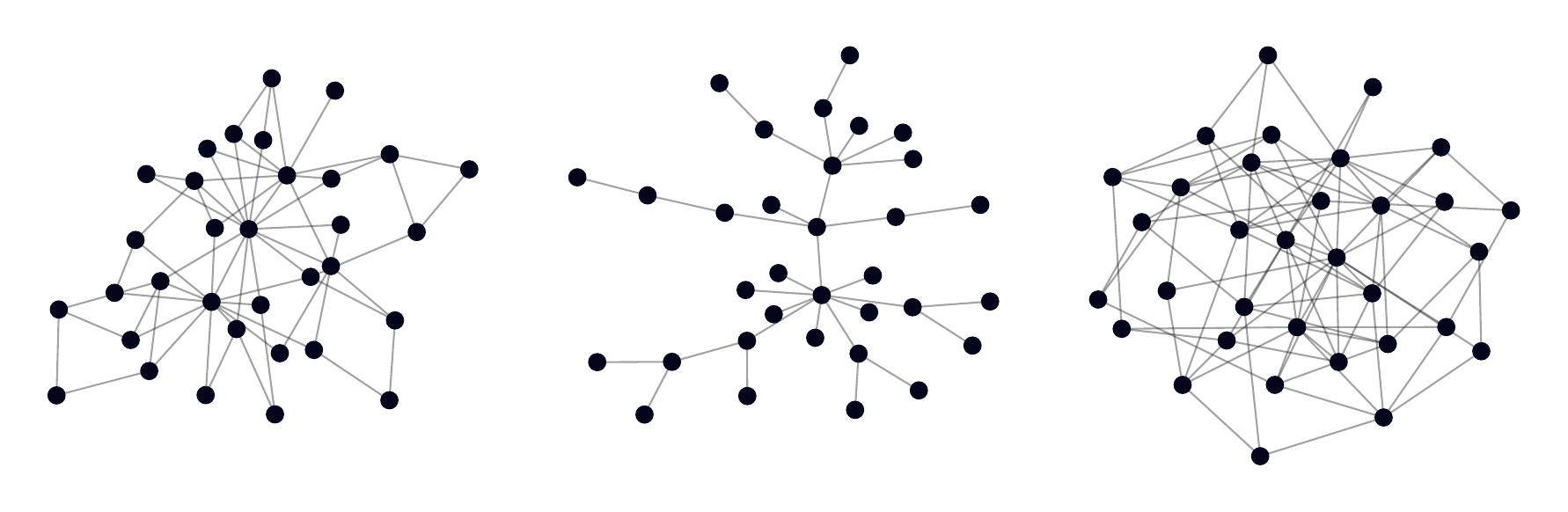}
  \caption{  
  \textbf{\barabasialbert{} topologies.} Each topology has $n$ = 33 nodes. 
  Left to right: topologies have preferential attachment parameter $p \in$ \{2, 1, 3\}. 
  Top to bottom: seeds $\in$ \{0, 1, 2\}. 
  }
\label{fig:ba_33}
\end{figure*}

We visualize all the topologies that we study in our experiments in Figs~\ref{fig:ba}--\ref{fig:ba_33}.

\clearpage
\newpage

\subsection{Impact of Topology}
\label{appendix:topo_impact}

\begin{figure*}[h]
  \includegraphics[width=\textwidth]{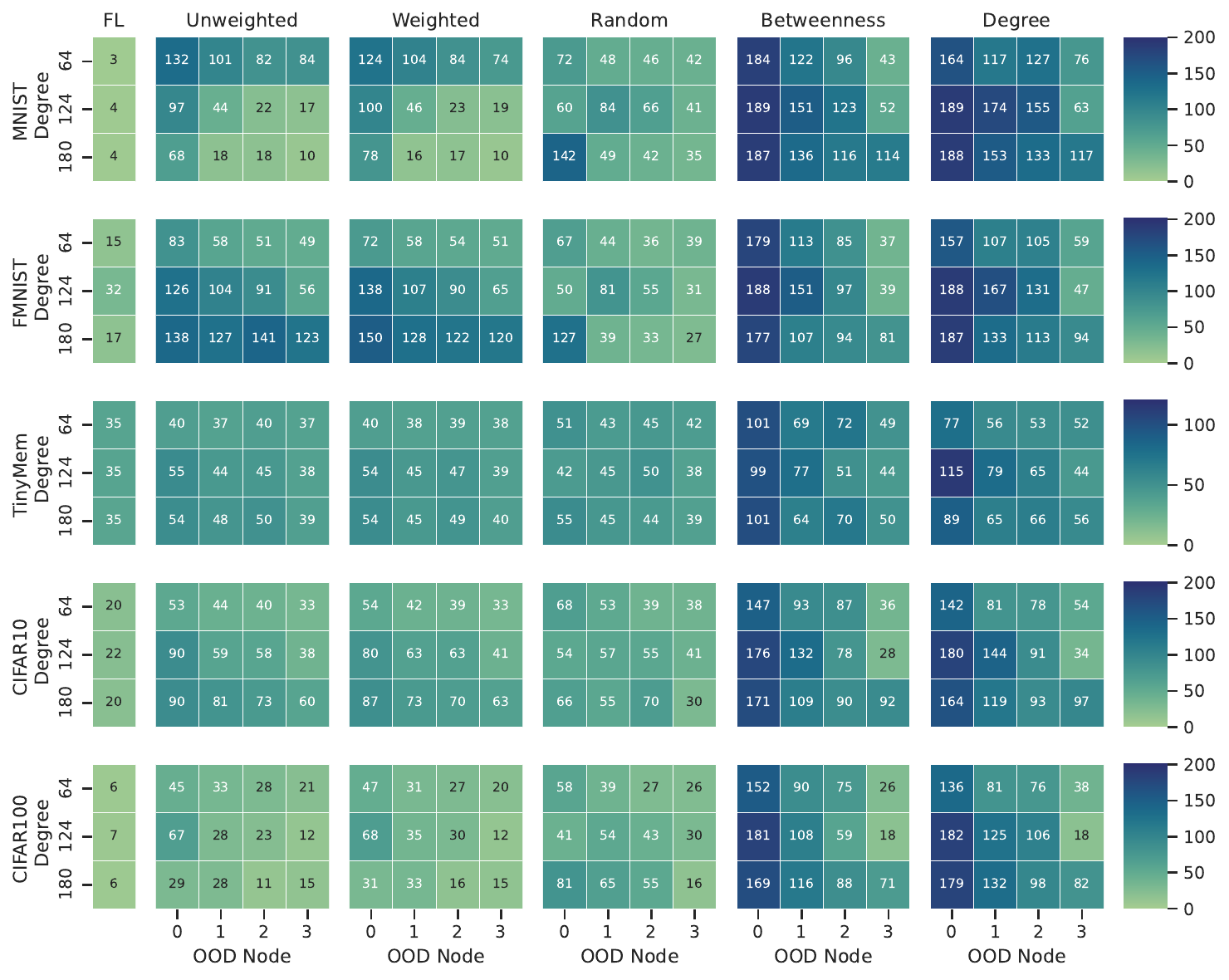}
  \caption{  
    \textbf{Impact of topology degree on aggregation strategy performance.} Experiments performed across \barabasialbert{} topologies with $n$ = 33 nodes and $p \in$ \{1, 2, 3\}. Higher $p$ means higher degree. Results averaged across three seeds. 
  }
\label{fig:degree_impact}
\end{figure*}

\begin{figure*}[h]
  \includegraphics[width=\textwidth]{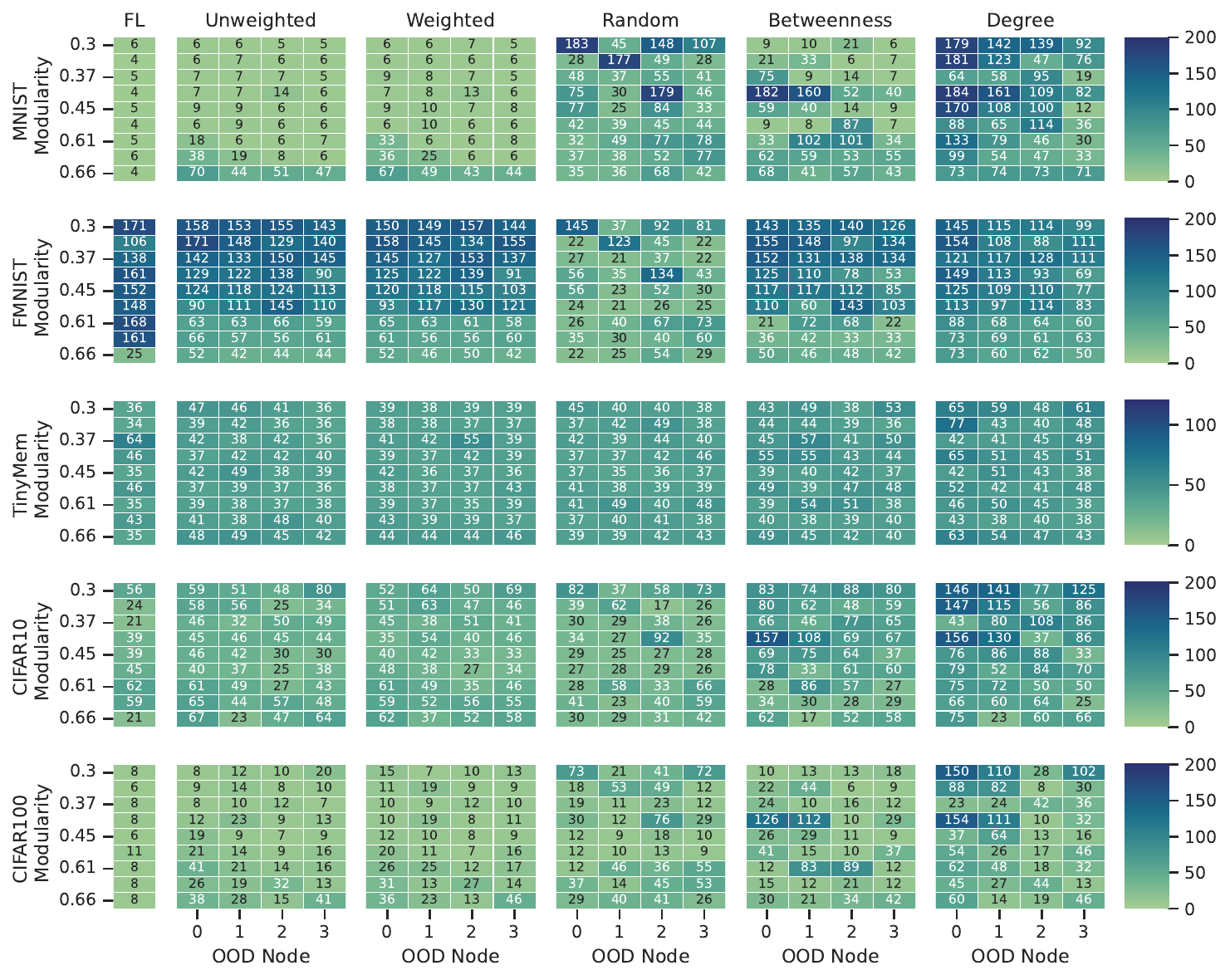}
  \caption{  
    \textbf{Impact of modularity on aggregation strategy performance.} Experiments performed across \stochasticblock{}{} topologies with $n$ = 33 nodes and three communities with varying levels of modularity. The probabilities of edges existing between communities $m_i$ to $m_j$ are $p_{i,j}$: if $i = j, p_{i_j}$ = 0.5, if $i \neq j$, then we varied $p_{i_j} \in$ \{0.009, 0.05, 0.9\}.
    Results shown for three seeds. (As a newly seeded \SB{} is generated, its \enquote{modularity} score changes from the previous seed; therefore, we do not average across seeds in this figure.)
  }
\label{fig:modularity_impact}
\end{figure*}

\begin{figure*}[h]
  \includegraphics[width=\textwidth]{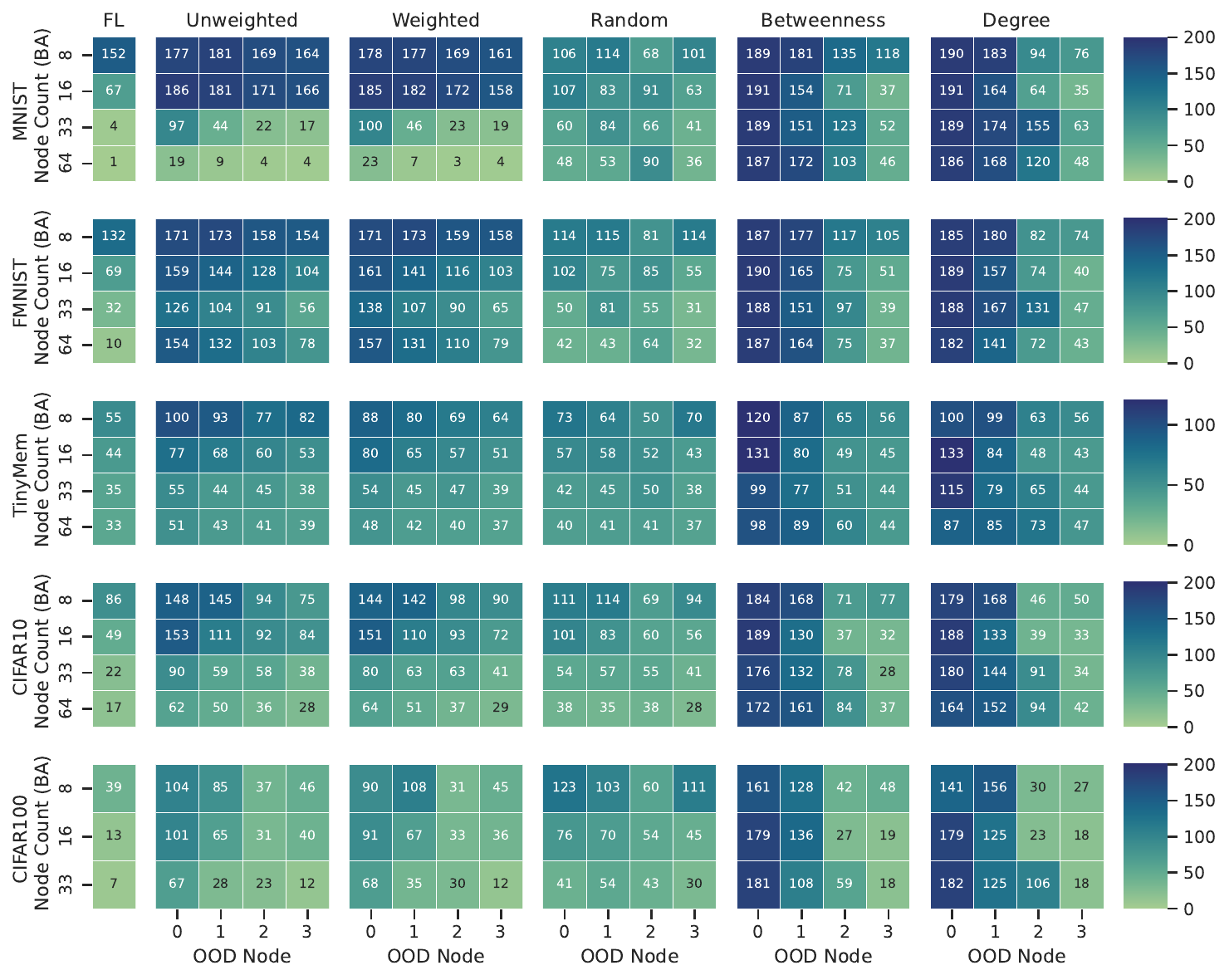}
  \caption{  
    \textbf{Impact of topology node count on aggregation strategy performance.} Experiments performed across \barabasialbert{} topologies with $p$ = 2, $n \in$ \{8, 16, 33, 64\} nodes. Results averaged across three seeds.
    We exclude experiment for CIFAR100 on \BA{} topologies w/ $n$ = 64 due to computational cost.
  }
\label{fig:nodes_ba_impact}
\end{figure*}

\begin{figure*}[h]
  \includegraphics[width=\textwidth]{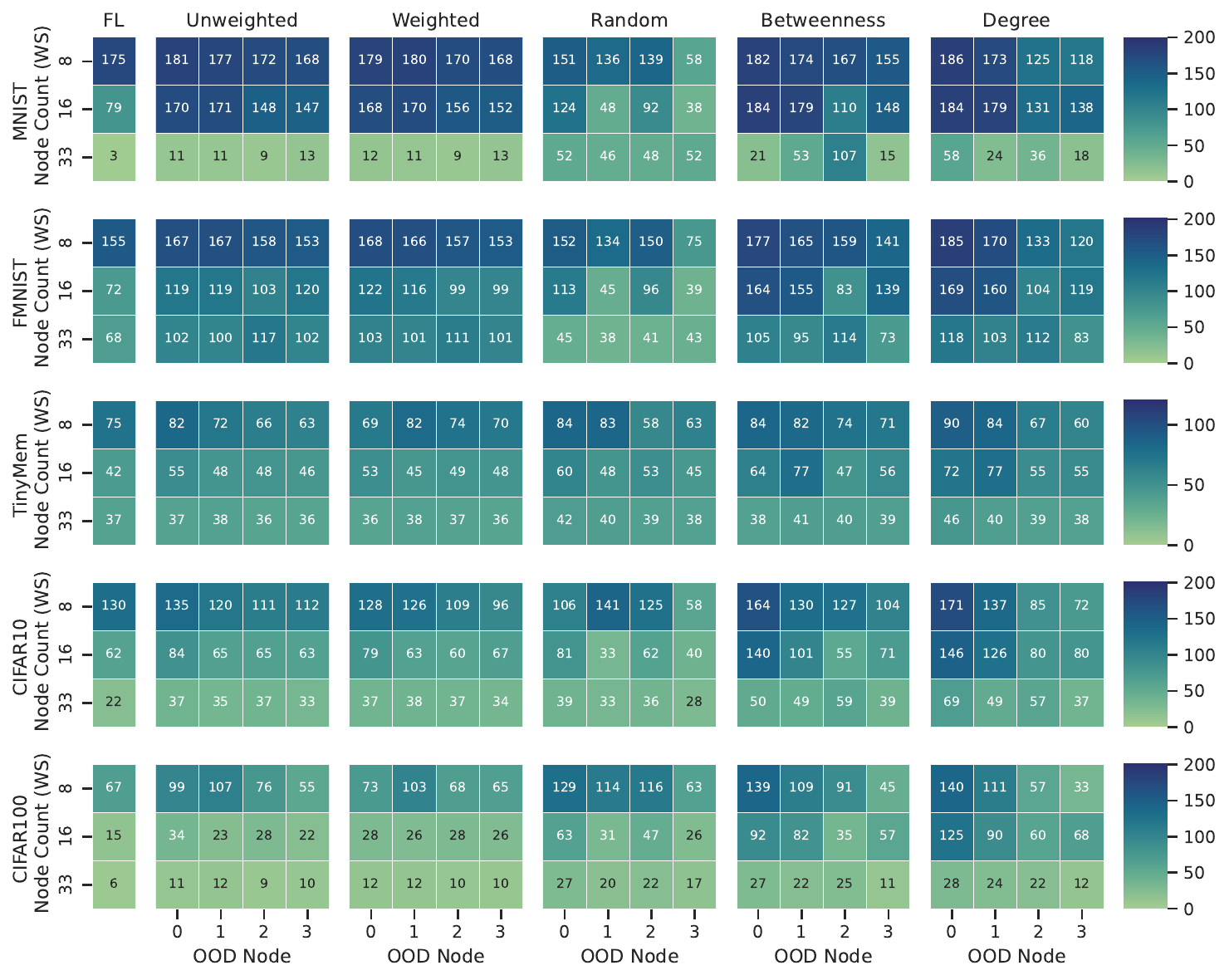}
  \caption{  
    \textbf{Impact of topology node count on aggregation strategy performance.} Experiments performed across \WS{}{} topologies with $k$ = 4, $u$ = 0.5, and $n\in$ \{8, 16, 33, 64\} nodes. Results averaged across three seeds. 
  }
\label{fig:nodes_ws_impact}
\end{figure*}

\begin{figure*}[h]
  \includegraphics[width=\linewidth]{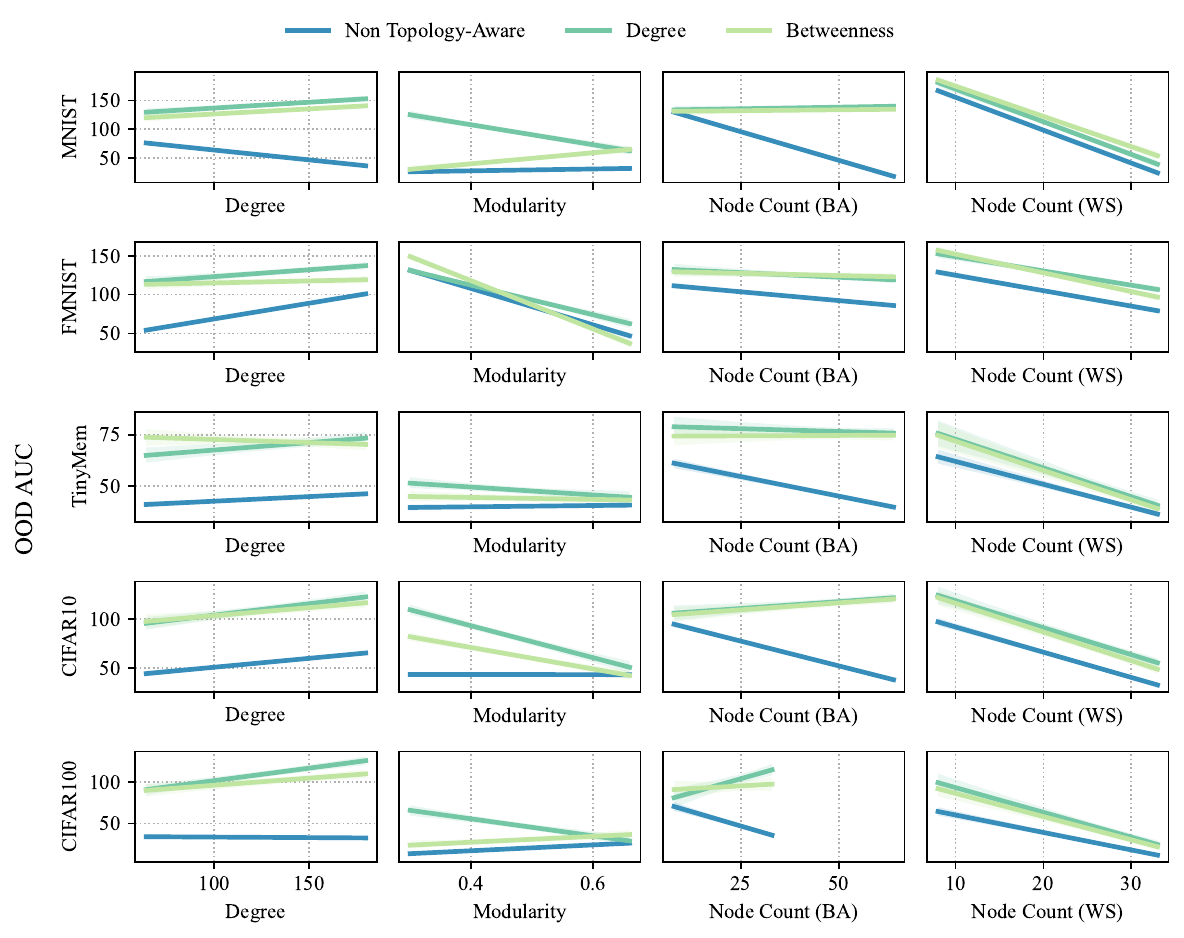}
  \caption{  
    \textbf{Impact of topology degree, modularity, node count on aggregation strategy performance.} From left to right: we plot the impact of topology degree, modularity, and node count on the OOD test accuracy AUC. Higher is better (indicates higher propagation of OOD knowledge).
    We exclude experiment for CIFAR100 on \BA{} topologies w/ $n$ = 64 due to computational cost.
  }
\label{fig:all_topo_impact}
\end{figure*}

\clearpage
\newpage

\subsection{\barabasialbert{} vs.\ \wattsstrogatz{} Degree Distributions}
\label{appendix:ba_vs_ws}

\begin{figure*}[h]
  \includegraphics[width=\textwidth]{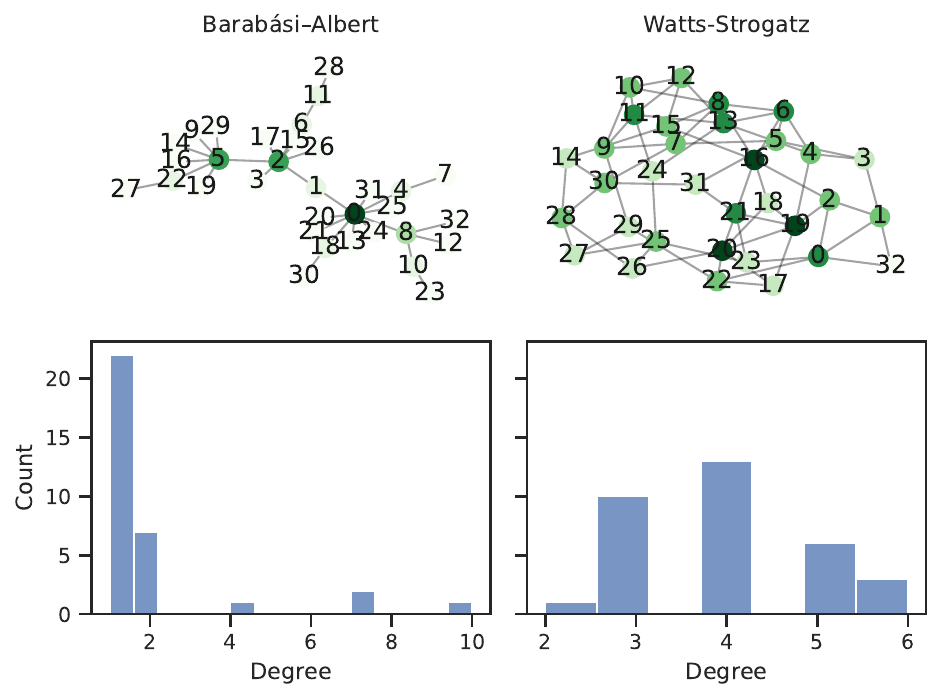}
  \caption{  
  \textbf{\barabasialbert{} vs.\ \wattsstrogatz{} degree distributions.} Histograms show the relative degree distributions show the relative degree distributions of both the \BA{} and \WS{}. Both topologies have $n$ = 33 nodes. \BA{} has $p$ = 1. \WS{} has $k$ = 4, $u$ = 0.5.
  \BA{} has a power-law degree distribution across its nodes while the \WS{} has a more normal degree distribution.
  Numbers on nodes indicate node identification number.
  Nodes colored by degree count (darker means higher degree).
  }
\label{fig:ba_vs_ws}
\end{figure*}

\clearpage
\newpage

\section{Computation Resources and Energy Usage}
\label{appendix:compute_energy_use}

\begin{table}[t]
    \caption{    
        \textbf{System parameters} for Aurora at ALCF~\citep{AuroraALCF} to estimate energy and carbon usage.
    }
    \centering
    \begin{tabular}{rccccccc}
    \toprule
         Machine & $c_\textrm{uf}$ & $c_\textrm{tdp}$  & $\textrm{ngpu}$ & $g_\textrm{uf}$ & $\mathit{gpu}_\mathit{tdp} $ & $\mathit{DRAM}$ & $\mathit{PUE}$ \\
    \midrule
        Aurora & 0.5 & 350 & 6 & 1 & 600 & 1024  & 1.58 \\
    
    \bottomrule
    \end{tabular}

    \label{tab:energy_carbon}
\end{table}
\begin{table}[t]
    \caption{    
        \textbf{Experiment counts} across topologies with varying device counts.
    }
    \centering
    \begin{tabular}{lcccc}
    \toprule
        Data & 8 Devices & 16 Devices & 33 Devices  & 64 Devices \\
    \midrule
        MNIST & 126 & 126 & 441 & 63  \\
        FMNIST & 126 & 126 & 441 & 63 \\
        TinyMem & 126 & 126 & 441 & 63 \\
        CIFAR10 & 126 & 126 & 441 & 63 \\
        CIFAR100 & 126 &  126 & 441 & n/a \\

    \bottomrule
    \end{tabular}

    \label{tab:experiment_counts}
\end{table}
\begin{table}[t]
    \caption{    
        \textbf{Energy and carbon estimates} for our experiments based on \autoref{eq:energy} \& \autoref{eq:carbon} respectively.
    }
    \centering
    \begin{tabular}{lcccc}
    \toprule
        Data & Machine & Node Hours & Energy (kWh)  & Carbon (Kg) \\
    \midrule
        MNIST & Aurora & 1229.12 & 8071.85 & 2421.55  \\
        FMNIST & Aurora & 1186.30 & 7790.65 & 2337.20  \\
        TinyMem & Aurora & 2336.25 & 15342.57 & 4602.77  \\
        CIFAR10 & Aurora & 3679.14 & 24161.54 & 7248.46 \\
        CIFAR100 & Aurora & 2255.33 & 4443.35 & 4443.35  \\
        \midrule
        All Experiments (total) & -- & 10686.14 & 70177.78 & 21053.33  \\

    \bottomrule
    \end{tabular}

    \label{tab:energy_carbon_estimates}
\end{table}
\begin{table}[t]
    \caption{    
        \textbf{Single Experiment Node hours.} Estimates from \BA{} topology with $p$ = 3 for 40 rounds of training. We exclude experiment for CIFAR100 with 64 device topologies due to computational constraints. All times tripled to account for extra debugging time, faulty runs, and any unaccounted for compute usage.
    }
    \centering
    \begin{tabular}{lcccc}
    \toprule
        Data & 8 Devices & 16 Devices & 33 Devices  & 64 Devices \\
    \midrule
        MNIST & 0.97 & 1.10 & 1.74 & 3.20 \\
        FMNIST & 0.76 & 1.07 & 1.71 & 3.19 \\
        TinyMem & 1.62 & 2.06 & 3.39 & 6.01 \\
        CIFAR10 & 2.33 & 3.15 & 5.37 & 9.84 \\
        CIFAR100 & 2.34 &  2.65 & 3.69 & n/a \\
    \bottomrule
    \end{tabular}

    \label{tab:single_experiment_estimates}
\end{table}

The experiments in this paper used approximately \textbf{70,178 kWh of energy, and 21,0153 Kg of carbon}.

We detail the computational resource (i.e., node hours, energy, carbon) used by our experiments below.

To calculate node hours, we calculate the average time all of our final experiments for two aggregation rounds across three trials. We then multiple by 20 for each experiment as we run each experiment for 40 aggregation rounds. We then \textit{triple that value} to account for extra debugging time, faulty runs, and any unaccounted for compute usage (see~\autoref{tab:single_experiment_estimates}). All experiments were run on the Aurora machine at Argonne Leadership Computing Facility \cite{AuroraALCF}.

To calculate energy usage and carbon cost of our experiments, we follow the methodology detailed by \citet{bouza2023estimate}:
\begin{equation}
    \mathit{energy} = 
    \mathit{NH} * 
    \left(
        \left(
            c_{\textrm{uf}} * c_{\textrm{tdp}}
        \right) 
        + 
        \left(
            ngpu * g_{\textrm{uf}} * gpu_{\textrm{tdp}}
        \right) 
        + 
        \left(
            0.3725~\textrm{W/Gb} * \mathit{DRAM}
        \right)
    \right) * \mathit{PUE},
\label{eq:energy}
\end{equation}
where \emph{NH} is node hours, $c_{\textrm{uf}}$ is the CPU usage factor, $c_{\textrm{tdp}}$ is the CPU's thermal design power,  \textit{ngpu} is the number of GPUs on a node, $g_{\textrm{uf}}$ is the GPU usage factor (we assume 100\% utilization), $g_{\textrm{tdp}}$ is the GPU's thermal design power, \textit{DRAM} is dynamic random access memory, and \textit{PUE} is the power usage efficiency. \textit{energy} is reported in watt hours. We record system-specific parameter values in \autoref{tab:energy_carbon}.

\begin{equation}
    \mathit{carbon} = 
    \left(
        \mathit{energy} / 1000
    \right) * CI
\label{eq:carbon}
\end{equation}

\noindent 

Above in \autoref{eq:carbon}, \emph{energy} is obtained in watt hours from~\autoref{eq:energy}, and \emph{CI} is the carbon intensity reported based on the yearly regional average for each computing center from \citep{electricityMap}. For Aurora's geographic location in 2025, \emph{CI} = 300 g/kWh. \emph{carbon} is reported in grams.

\clearpage
\newpage

\end{document}